\documentclass[final]{cvpr}
\newcommand{\final}{0}
\usepackage{times}
\usepackage{epsfig}
\usepackage{graphicx}
\usepackage{amsmath}
\usepackage{amssymb}
\usepackage{multirow}
\usepackage{booktabs}
\usepackage{amssymb}
\usepackage{url}
\usepackage{ifthen}
\usepackage{color}
\definecolor{XingjiaColor}{rgb}{0.0,0.1,0.9}
\definecolor{FanColor}{rgb}{0.8,0,0.8}
\definecolor{JunColor}{rgb}{0, 0, 0}
\newcommand{\xingjia}[1]{{\color{XingjiaColor}[Xingjia: #1]}}
\newcommand{\fan}[1]{{\color{FanColor}[Fan: #1]}}
\newcommand{\jun}[1]{{\color{JunColor}#1}}

\newcommand{\warning}[1]{{\it\color{red} #1}}
\newcommand{\toremove}[1]{{\it\color{red} (To remove) #1}}
\newcommand{\note}[1]{{\it\color{blue} #1}}

\newcommand{\nothing}[1]{}
\ifthenelse{\equal{\final}{1}}
{
 
\renewcommand{\fan}[1]{}
\renewcommand{\xingjia}[1]{}
\renewcommand{\jun}[1]{}

\renewcommand{\warning}[1]{}
\renewcommand{\toremove}[1]{}
\renewcommand{\note}[1]{}
\renewcommand{\nothing}[1]{}
}{}

\usepackage[pagebackref=true,breaklinks=true,colorlinks,bookmarks=false]{hyperref}

\def\cvprPaperID{3784} % *** Enter the CVPR Paper ID here
\def\confYear{CVPR 2022}
\newcommand{\figtext}[1]{{\footnotesize #1}}

\begin{document}

%%%%%%%%% TITLE
\title{SIOD: Single Instance Annotated Per Category Per Image for Object Detection }
\author{Hanjun Li\textsuperscript{1}\thanks{Work partially done during the Youtu Lab internship},
Xingjia Pan\textsuperscript{2},
Ke Yan \textsuperscript{2}\thanks{Corresponding author},
Fan Tang \textsuperscript{3},
Wei-Shi Zheng\textsuperscript{1,4,5} \footnotemark[2]
\\
\textsuperscript{1}{School of Computer Science and Engineering, Sun Yat-sen University}\\
\textsuperscript{2}{Youtu Lab, Tencent} \quad
\textsuperscript{3}{Jilin University} \quad 
\textsuperscript{4}{Peng Cheng Laboratory}\\
\textsuperscript{5}{Key Laboratory of Machine Intelligence and Advanced Computing, Ministry of Education}\\
\tt\small lihj85@mail2.sysu.edu.cn, \{xjia.pan,tfan.108\}@gmail.com,kerwinyan@tencent.com, wszheng@ieee.org
}
% \and

\maketitle
% \thispagestyle{empty}

%%%%%%%%% ABSTRACT
% \input{Sections/0_abstract_bak}
\begin{abstract}
%Object detection under imperfect data receives great attention recently. Weakly supervised object detection (WSOD) suffers from severe localization issues due to the lack of instance-level annotation, while semi-supervised object detection (SSOD) remains challenging led by the inter-image discrepancy between labeled and unlabeled data. In this study, we propose the Single Instance annotated Object Detection (SIOD), requiring only one instance annotation for each existing category in an image. Degraded from inter-task (WSOD) or inter-image (SSOD) discrepancies to the intra-image discrepancy, SIOD provides more reliable and rich prior knowledge for mining the rest of unlabeled instances and trades off the annotating cost and performance. Under SIOD setup, we propose a simple yet effective framework, termed Dual-Mining (DMiner), which includes a Similarity-based Pseudo Label Generating module (SPLG) and a Pixel-level Group Contrastive Learning module (PGCL). SPLG firstly mines latent instances from feature representation space to alleviate the annotation missing problem. To avoid being misled by inaccurate pseudo labels, we propose PGCL to boost the tolerance to false pseudo labels. Extensive experiments on MS COCO verify the superiority of SIOD setup and the proposed method obtains consistent and significant improvements compared to baseline methods and achieves comparable results with fully supervised object detection (FSOD) methods with only 40$\%$ instances annotated. 
Object detection under imperfect data receives great attention recently. Weakly supervised object detection (WSOD) suffers from severe localization issues due to the lack of instance-level annotation, while semi-supervised object detection (SSOD) remains challenging led by the inter-image discrepancy between labeled and unlabeled data. In this study, we propose the Single Instance annotated Object Detection (SIOD), requiring only one instance annotation for each existing category in an image. Degraded from inter-task (WSOD) or inter-image (SSOD) discrepancies to the intra-image discrepancy, SIOD provides more reliable and rich prior knowledge for mining the rest of unlabeled instances and trades off the annotation cost and performance. Under the SIOD setting, we propose a simple yet effective framework, termed Dual-Mining (DMiner), which consists of a Similarity-based Pseudo Label Generating module (SPLG) and a Pixel-level Group Contrastive Learning module (PGCL). SPLG firstly mines latent instances from feature representation space to alleviate the annotation missing problem. To avoid being misled by inaccurate pseudo labels, we propose PGCL to boost the tolerance to false pseudo labels. Extensive experiments on MS COCO verify the feasibility of the SIOD setting and the superiority of the proposed method, which obtains consistent and significant improvements compared to baseline methods and achieves comparable results with fully supervised object detection (FSOD) methods with only 40\% instances annotated.
Code is available at \url{https://github.com/solicucu/SIOD}.
\end{abstract}
% \input{Figures/base_keep1_ap}
%%%%%%%%% BODY TEXT
% \input{Sections/1_introduction_bak}
\section{Introduction}
\label{sec:intro}
% bring out the theme 
With the boom of Convolutional Neural Network (CNN) and Vision Transformer \cite{dosovitskiy2020image,krizhevsky2012imagenet,he2016deep,liu2021swin,simonyan2014very,tan2019efficientnet}, object detection~\cite{bochkovskiy2020yolov4,carion2020end,lin2017feature,liu2016ssd,redmon2016you,redmon2017yolo9000,ren2015faster,zhou2019objects} has achieved great improvements with a large number of instance-level annotations.
However, these annotations are not only labor-intensive and time-consuming, but also prevent detectors from generalizing to most realistic scenarios where only few labeled data is available. 
% \begin{figure}
%     \begin{center}
%         \includegraphics[width=\linewidth]{Images/det_types.pdf}
%     \end{center}
%     \caption{Different annotation setup for (a) WSOD, (b)SSOD, (c) SAOD, and our (d) SIOD. }
%     \label{fig:det_types}
% \end{figure}

\newcommand\dettypefigurewidth{0.49}
\begin{figure}
\centering

    \includegraphics[width=\dettypefigurewidth\linewidth]{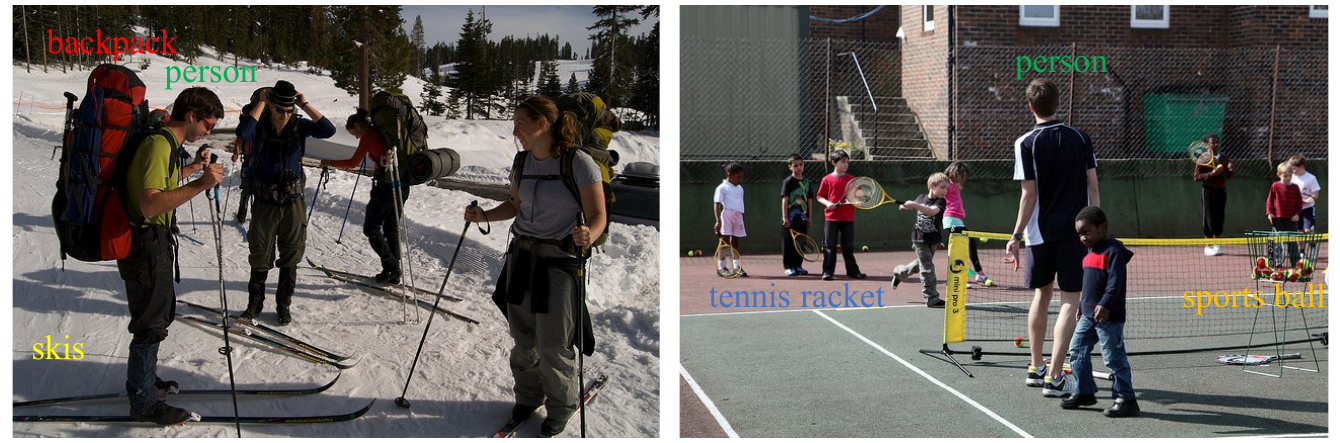}
    \includegraphics[width=\dettypefigurewidth\linewidth]{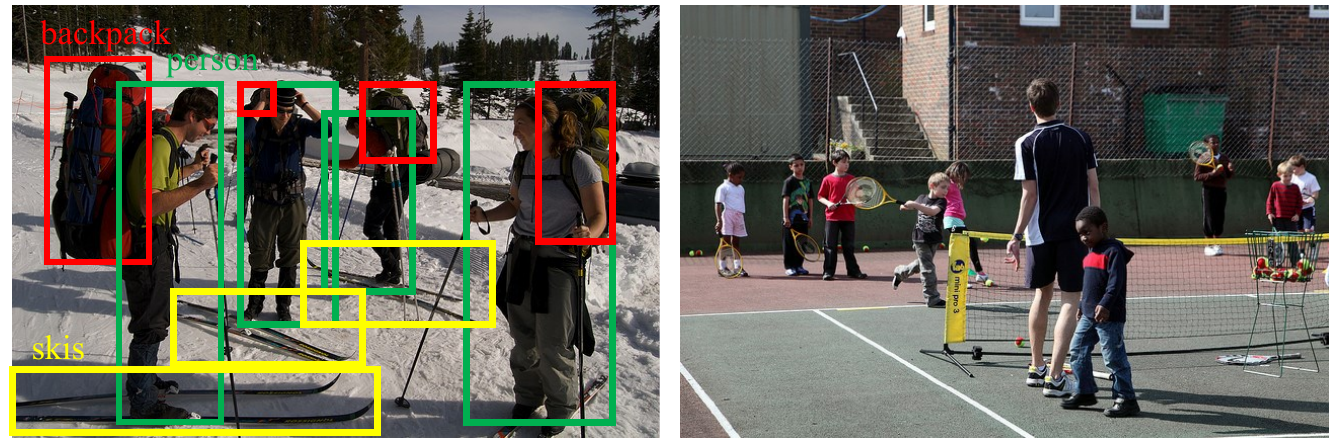} \\
    \vspace{-0.5mm}
     \begin{minipage}{\dettypefigurewidth\linewidth}
    \centering
    \figtext{(a) WSOD}
    \end{minipage}
    \begin{minipage}{\dettypefigurewidth\linewidth}
    \centering
    \figtext{(b) SSOD}
    \end{minipage} \\
    \vspace{0.5mm}
    \includegraphics[width=\dettypefigurewidth\linewidth]{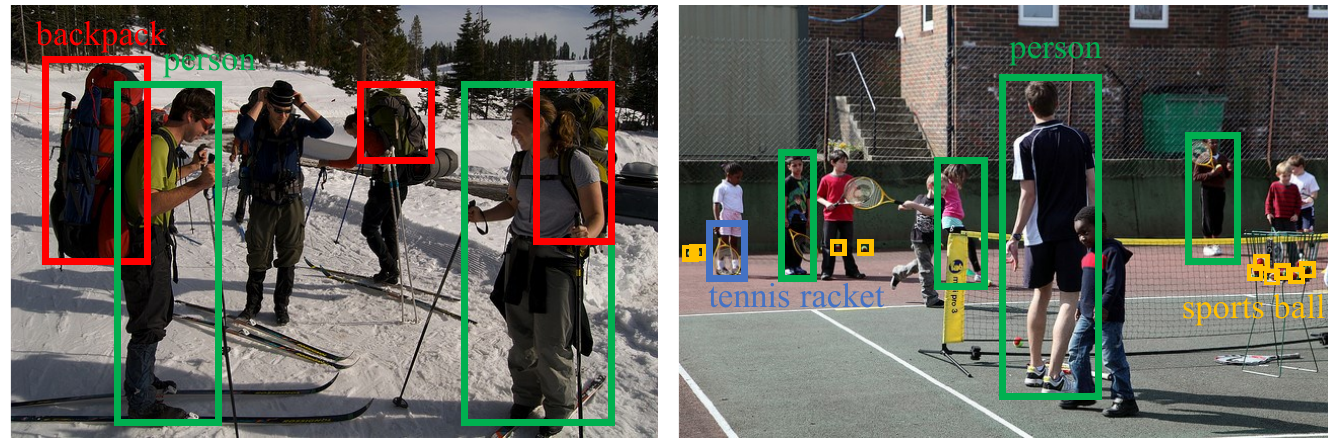}
    \includegraphics[width=\dettypefigurewidth\linewidth]{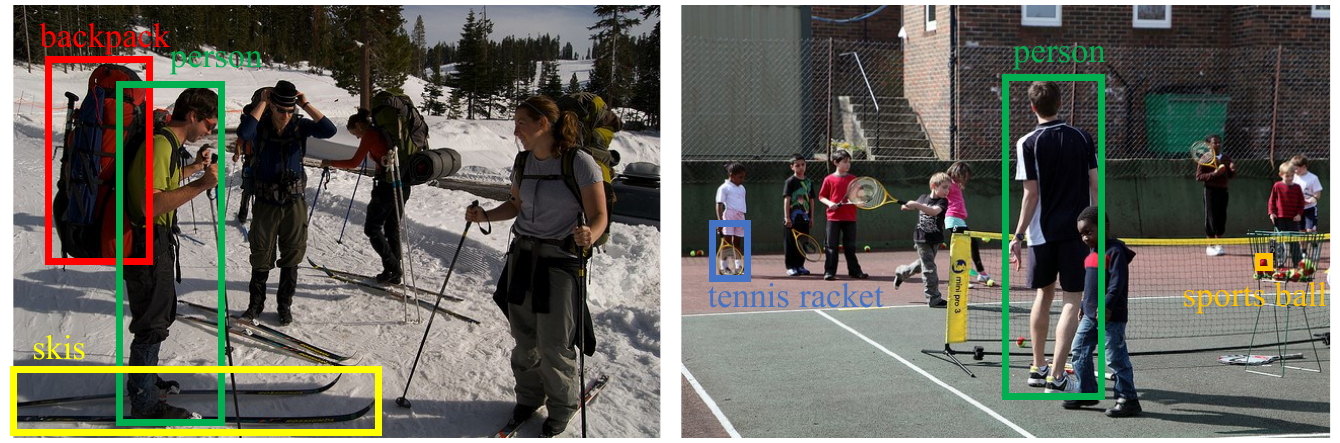} \\
    \vspace{-0.5mm}
    \begin{minipage}{\dettypefigurewidth\linewidth}
    \centering
    \figtext{(c) SAOD}
    \end{minipage}
    \begin{minipage}{\dettypefigurewidth\linewidth}
    \centering
    \figtext{(d) SIOD (Ours)}
    \end{minipage}
    \caption{Different annotation setup for (a) WSOD, (b) SSOD, (c) SAOD and (d) SIOD (Ours).}
    \label{fig:det_types}
\end{figure}

% defect of WSOD 
Weakly supervised object detection (WSOD), which requires only image-level labels for training, has received much attention in computer vision community.
Although great advances~\cite{bilen2016weakly,lin2020object,tang2017multiple,shen2020uwsod,dong2021boosting} have been achieved in recent years, it still remains a huge performance gap between WSOD and FSOD.
WSOD suffers from severe localization issues due to the large discrepancy between image-level annotation and instance-level task. 
% defect of SSOD 
Semi-supervised object detection (SSOD) is an alternative few-label object detection task, where only a small number of instance-level annotations are available. 
SSOD methods~\cite{jeong2019consistency,jeong2021interpolation,bachman2014learning,MeanTeacher,FeatMatch,zhou2021instant} obtain superior localization accuracy compared to WSOD methods, but the inter-image discrepancy between labeled and unlabeled data limits further improvement by a large margin due to the lack of explicit inter-image communication in popular mini-batch optimization manner.

% defect of sparse annotation methods
Sparsely annotated object detection (SAOD) is recently proposed, which annotates a part of instances in each image.
Specifically, SAOD methods~\cite{wang2020co,xu2019missing,zhang2020solving} usually imitate sparse annotation by randomly erasing different proportion of annotations from completely annotated object detection datasets~\cite{lin2014microsoft}.
In this way, it's inevitable that there are not any annotations for all instances of some categories in an image, which leads to inter-image discrepancy as SSOD.\nothing{ and brings a huge challenge for accurate object detection.} 
To this end, we propose a new task, termed as \emph{S}ingle \emph{I}nstance annotated \emph{O}bject \emph{D}etection (\emph{\textbf{SIOD}}), which annotates only one instance for each existing category in an image.
Compared to WSOD, SSOD and SAOD, SIOD reduces the inter-task or inter-image discrepancies to intra-image discrepancy and trades off the annotating cost and performance.
Fig.~\ref{fig:det_types} illustrates the annotation details of WSOD, SSOD, SAOD and the proposed SIOD.

Pseudo label-based methods are the most popular solution under the imperfect data and achieve impressive progress.
However, pseudo labels generated by detector in the early training phase are usually inaccurate and make it difficult for stable training.
\nothing{Additionally, the inaccurate pseudo labels would amplify the error between the pseudo labels and trained detector.}
In this study, we propose a simple yet effective framework under the SIOD setup, called Dual-Mining (DMiner), which consists of a Similarity-based Pseudo Label Generating module (SPLG) and a Pixel-level Group Contrastive Learning module (PGCL).
In contrast to detector-based pseudo labels, the SPLG instead utilizes the feature similarity to mines latent instances, which is based on the ability of equi-variance of CNN.
%这个To..., propose .... to, 这个句式读起来感觉很怪，为了什么，提出了什么来达到什么，成了提出了一个东西达成两种目的的感觉，会给人一种感觉，这个东西到底是为了实现哪个目标？
\nothing{To avoid being misled by inaccurate pseudo labels, we propose the PGCL to boost the tolerance to false labels with the help of group contrastive learning to self-mine a group of positive pairs for each category and minimize the distances between instances of same category in each image.}
\jun{We then propose the PGCL, which self-mines a group of positive pairs for each category for group contrastive learning, to boost the tolerance to false pseudo labels and minimize the distances between instances of same category in each image.}
% % protocol
COCO style evaluation protocol does not filter the detected boxes with extremely low confidence and thus results in illusory advances in object detection by recalling a large number of objects with low confidence.
We therefore introduce additional confidence constraint to coco style evaluation metrics that a predicted box is determined as a true match only when it satisfies the specific IoU (Interaction Over Union) and confidence threshold. 
% results
\nothing{We conduct extensive experiments on MS COCO~\cite{lin2014microsoft} across several architectures. 
Our method obtains consistent and significant improvements compared to baseline methods and achieves comparable results with methods under FSOD setting with only 40$\%$ labeled data.}

% contributions
In summary, the contributions in this study include:
\begin{itemize}
    \item Investigate the SIOD task, which provides more possibility of the development of object detection with lower annotated cost.
    \item Propose the DMiner framework to mine unlabeled instances and boost the tolerance to false pseudo labels.  
    \item Extensive experiments verify the superiority of SIOD and the proposed DMiner obtains consistent and significant gains compared with baseline methods. 
\end{itemize}

\section{Related Works}
\subsection{Object Detection}
\noindent\textbf{Object detection under full supervision.}
Fully supervised object detection has achieved great progress in recent years. 
Most modern detection methods can be roughly divided into two-stage or one-stage methods.
Two-stage methods~\cite{ren2015faster,lin2017feature,cai2019cascade,guo2021distilling} usually first generate high-quality proposals by introducing Region Proposal Networks (RPN) and then apply a refine stage to obtain final predictions.
Meanwhile, one-stage methods~\cite{liu2016ssd,redmon2017yolo9000,bochkovskiy2020yolov4,zhou2019objects,tian2019fcos} directly regress bounding boxes and predict class probabilities, which lead to high efficiency.
After CornerNet~\cite{law2018cornernet}, several methods~\cite{zhou2019objects,zhu2020deformable, sun2021sparse, yang2019reppoints, carion2020end, zhang2021varifocalnet} are proposed to remove anchor setting by directly predicting absolute bounding box \emph{w.r.t} input image and most of these methods follow the setup of one-stage pipeline. 
%
%Recently, transformer equipped methods \cite{carion2020end,zhu2020deformable, dai2021up} receive great attention, which present a end-to-end pipeline.
%
%Detection Transformer (DETR)~\cite{carion2020end} adapts the Transformer encoder-decoder head built upon CNN backbone and presents an end-to-end optimization objective for set prediction without non-maximum suppression (NMS).Several variants, such as Deformable DETR~\cite{zhu2020deformable} and UP-DETR~\cite{dai2021up} are proposed to improve its training efficiency and performance.
% The transformer based methods\cite{carion2020end,zhu2020deformable, dai2021up} further simplify the pipeline of object detection.
% In the past few years, both anchor-based\cite{ren2015faster,lin2017feature, tian2019fcos, bochkovskiy2020yolov4, chen2021you,guo2021distilling} and anchor-free\cite{zhu2020deformable, sun2021sparse, law2018cornernet, yang2019reppoints, carion2020end, zhang2021varifocalnet} object detection have achieved remarkable progress thanks to the development of CNN and Transformer\cite{vaswani2017attention,liu2021swin,graham2021levit}.
However, the great success of these approaches heavily depend on a large number of instance-level annotations, which is notoriously labor-intensive and time-consuming. 

\noindent\textbf{Object detection under imperfect data.}
Weakly and semi-supervised object detection attract increasing attention because they relieve the demand of expensive instance-level annotations.
Most of WSOD methods~\cite{bilen2016weakly,lin2020object,tang2017multiple,shen2020uwsod,pan2021unveiling,gao2021token,zhang2018adversarial} adopt multiple instance learning (MIL)~\cite{dietterich1997solving} or CAM~\cite{zhou2016learning} to mine the latent object proposals with image-level labels followed by some instance refinement modules.
Due to the lack of instance-level annotations, there is still a large performance gap between WSOD and FSOD.
SSOD approaches leverage both limited instance-level labeled data and large amounts of unlabeled data.
Consistency regularization based methods~\cite{jeong2019consistency,jeong2021interpolation} help to train a model to be robust to given perturbed inputs, but the effect is limited and heavily depends on the data augmentation strategies~\cite{ma2022mitigating}.
Pseudo-label based methods~\cite{bachman2014learning,MeanTeacher,FeatMatch} are the current state-of-the-arts, and most of them conduct a complex multi-stage training schema: generating pseudo labels and re-training process. 
The final performance is limited by the quality of pseudo labels.
Recently, SAOD methods~\cite{wang2020co,xu2019missing,zhang2020solving} randomly annotate a proportion of instances in each image, which inevitably lead to imbalance for each category.
Different from aforementioned setup, SIOD annotates one instance for each existing category in each image to achieve the purpose of reducing annotation and retaining rich information.
\begin{figure*}[!t]
    \begin{center}
        \includegraphics[width=0.99\linewidth]{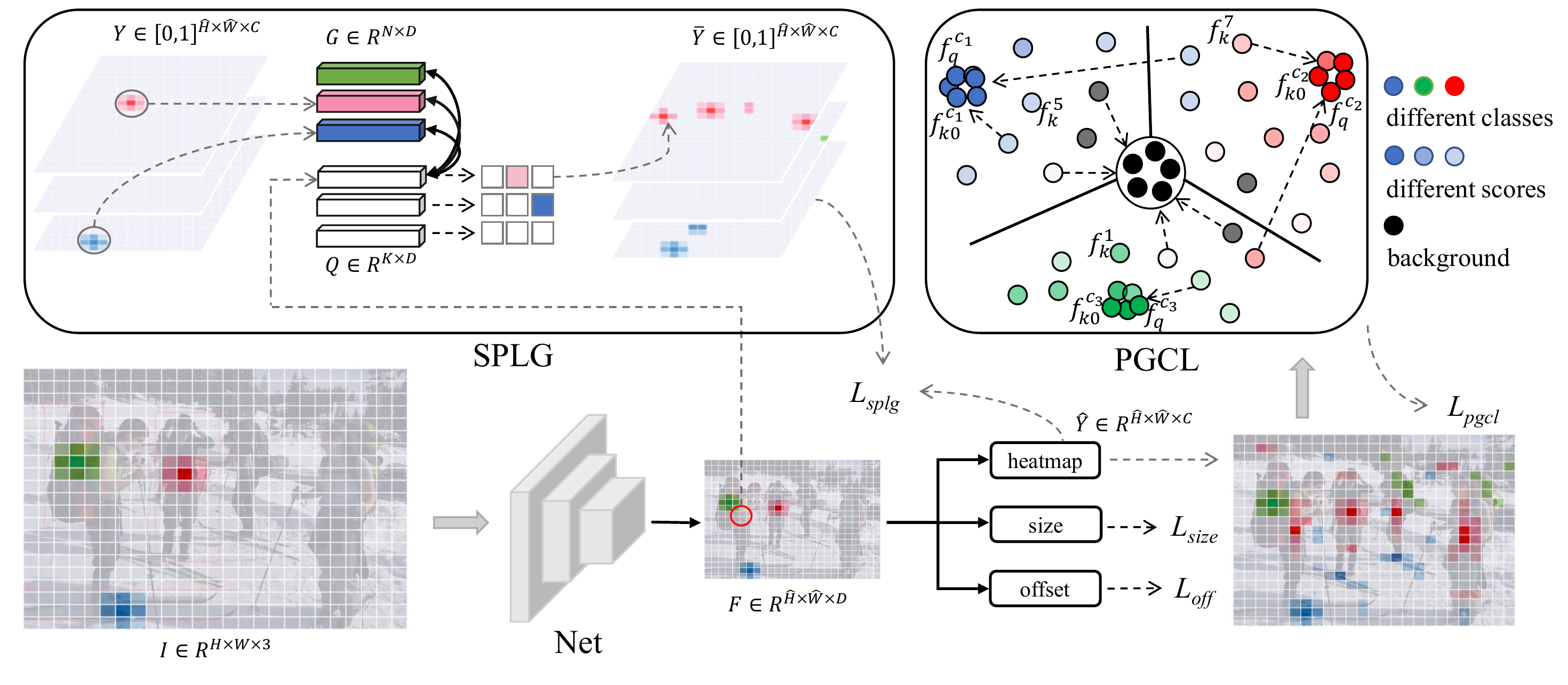}
    \end{center}
    \caption{The overview of proposed DMiner, which consists of a Similarity-based Pseudo Label Generating module (SPLG) and a Pixel-level Group Contrastive Learning module (PGCL). \jun{Let $M_{15}, M_{27}, M_{31}$=1 and refer to Eq.~(\ref{eq:loss_pgcl}) to understand the meaning of $M$.}}
    \label{fig:main_framework}
\end{figure*}
\subsection{Contrastive Learning}
Contrastive learning~\cite{he2020momentum, chen2020improved,zhu2021improving, wang2021self} has been widely used in the filed of self-supervised learning. 
Most of these methods use instance discrimination~\cite{wu2018unsupervised} as pretext task to pretrain a network and then fine-tune for different downstream tasks (\eg classification, object detection and segmentation). 
Contrastive learning aims to minimize the distance between the positive pairs, namely two different augmented views of same image, and push away negative pairs. 
Specifically, Xie \etal~\cite{xie2021detco} introduces contrastive learning between global image and local patches at multi-level features for pretraining and then transfer the learned model to object detection task.
Xie \etal~\cite{xie2021propagate} introduces pixel-level pretext tasks for learning dense feature representations which are friendly to dense prediction tasks (\eg object detection and segmentation).
To mitigate the reliance on pseudo-labels and boost the tolerance to false pseudo labels, Wang \etal~\cite{wang2021self} proposes a Pseudo Group Contrast mechanism (PGC) to address the challenge of confirmation bias in self-training.
However, all of them need to maintain a momentum encoder for extracting key features and a large feature queue, which is relatively cumbersome and resource-consuming.
% However, all of them need to maintain a large feature queue (memory bank) and a momentum encoder for extracting key features, which is relative cumbersome and resource-consuming.
To alleviate the error amplification issue led by inaccurate pseudo labels and mine more unlabeled instances in object detection, we design a pixel-level group contrastive learning module (PGCL). \nothing{to explore the intrinsic structure of data. }Note that PGCL is applied to each image independently without extra momentum encoder. 
% It is more convenient and efficient compared with general Contrastive Learning methods.

% 
% Different from general Contrastive Learning, it maintains a queue for each category. 
% and the feature of an image is pushed in the queue of class c when it has ground truth class label c or its pseudo class label is class c.
% However, all of them need to maintain a large feature queue (memory bank) and a momentum encoder for extracting key features, which is relative cumbersome and resource-consuming. 
% In this paper, we propose a pixel-level group contrastive learning in each image for object detection. Since each image may have a number of instances, we can easily construct a batch of positional features from feature map without extra momentum encoder according to the prediction. In this way, we can minimize the distance between the unlabeled instances and reference instance of the same class thus promotes the PCL(introduced in \ref{sec:pseudolabel}) for mining more latent instances for training. 
\section{DMiner}
\subsection{Overview}
In this paper, we adopt CenterNet~\cite{zhou2019objects} as our basic framework. 
%CenterNet is an end-to-end, simple, and anchor-free detector, which directly predicts whether each position is a central point of an instance and corresponding height and width. 
%Specifically, it consists of a backbone (\eg ResNet \cite{he2016deep}, DLA \cite{yu2018deep}, Hourglass \cite{newell2016stacked}) and three separate head branches for different predictions. 
Let $I \in \mathbb{R}^{H \times W \times 3}$ be an input image of height H and width W. 
Given a backbone $Net$, it firstly extracts the feature map $F \in \mathbb{R}^{\hat{H} \times \hat{W} \times D} = Net(I)$, where $\hat{H} = \frac{H}{s}$, $\hat{W}=\frac{W}{s}$, and $s$ is the downsample stride \emph{w.r.t} the input. 
$D$ is the feature dimension.
The features $F$ are then fed into a classifier head for predicting the category heatmap $\hat{Y} \in \mathbb{R}^{\hat{H} \times \hat{W} \times C}$, where $C$ is the number of category. 
Given an instance ground truth annotation $(cp_x,cp_y,w,h,c)$, where $(cp_x,cp_y)$ denotes the coordinates of the center point of the instance and $w,h,c$ are the width, height and category respectively, we generate a target category heatmap $Y \in [0,1]^{\hat{H} \times \hat{W} \times C}$ with Gaussian kernel function in Eq.~(\ref{eq:gaussian}), where $\hat{p}_x=\lfloor \frac{cp_x}{s} \rfloor$, $\hat{p}_y=\lfloor \frac{cp_y}{s} \rfloor$ and $\sigma_{wh}$ is an object size-adaptive standard deviation~\cite{law2018cornernet}.
\begin{equation}
    Y_{yxc}=exp(-\frac{(x-\hat{p}_x)^2+(y-\hat{p}_y)^2}{2\sigma_{wh}^2}).
    \label{eq:gaussian}
\end{equation}
Under the SIOD setup, directly assigning all the unlabeled region as background undoubtedly hurts the training process and deteriorates detector performance.  
% From the target $Y$, it is observed that those unlabeled instances are treated as background, which undoubtedly hurts the training process and deteriorates detector performance. 
To alleviate the annotation missing issue, we propose the Dual-Mining (DMiner) framework as shown in Fig.~\ref{fig:main_framework}, which consists of a Similarity-based Pseudo Label Generating module (SPLG) and a Pixel-level Group Contrastive Learning module (PGCL). 
The SPLG recalls several latent instances based on feature similarity between the reference instances (labeled) and the rest of the unlabeled region.
The model only utilizing the pseudo labels generated by SPLG is easily confused by false pseudo labels since it focuses on learning a hyperplane for discriminating each class from the other classes~\cite{wang2021self}.
Therefore, we further design the PGCL module to boost the tolerance to false pseudo labels, which is inspired from that contrastive learning loss focuses on exploring the intrinsic structure of data and is naturally independent of false pseudo labels~\cite{wang2021self}.
The overall training objective is as follows: 
\begin{equation}
%    L_{det}=L_{cls} + \lambda_{off}L_{off} + \lambda_{size}L_{size}
     L_{total} = L_{splg} + \lambda_{pgcl}L_{pgcl} +  \lambda_{off}L_{off} + \lambda_{size}L_{size}
\end{equation}
where $L_{off}$ and $L_{size}$ are center point offset and size regression losses following CenterNet~\cite{zhou2019objects}, $L_{splg}$ is the modified focal loss of the SPLG module (in Sec.~\ref{sec:pseudolabel}) and $L_{pgcl}$ is the loss of PGCL module (in Sec.~\ref{sec:ContrastiveLearning}). 
$\lambda_{pgcl}, \lambda_{off}, \lambda_{size}$ are the weight parameters of $L_{pgcl}, L_{off}, L_{size}$ losses, respectively.

\subsection{Similarity-based Pseudo Label Generating}
\label{sec:pseudolabel}
Under SIOD setup, we can obtain a labeled reference instance for each existing category in an image.
To solve the annotation missing problem, we propose to recall the unlabeled instances according to the feature similarity between the labeled reference instances and the rest of unlabeled data.  
% Our main target is to find out the other unlabeled instances and relabel them instead of treating corresponding position as background. The first intuitive scheme is that we can relabel each position according to the similarity between positional feature vector and the feature vector of reference instances.
% Specifically, we first normalize the feature map $F$ to $\hat{F}$ using L2-normalization. 
Let $C_I=\{c_1,c_2,..,c_N\}$ denotes the existing $N$ categories in current image $I$.
We can easily obtain the feature vector of each reference instance via Eq.~(\ref{eq:gfeat}).
\begin{equation}
    G_{c_i} = l_2(\sum_{yx} Y_{yxc_i} \hat{F}_{yx}) 
    \label{eq:gfeat}
\end{equation}
where $l_2$ denotes the L2-normalization. 
Let $\hat{P_U}=\{\hat{p_i}|$ unlabeled at position $\hat{p_i} \}$ denotes the unlabeled pixels in the feature map $F$ and $Q \in \mathbb{R}^{ K \times D}$ denotes the feature vectors of unlabeled data, where $K=|\hat{P_U}|$ indicates the number of unlabeled pixels.
We then attain the cosine similarity $S = QG^T$ between the reference instances and the rest of unlabeled data via a dot-product operation, where $S \in [0,1]^{K \times N}$.
According to the similarity matrix $S$, we can construct a pseudo category heatmap $\widetilde{Y} \in [0,1]^{\hat{H} \times \hat{W} \times C}$. 
For each position $\hat{p_i}$, $S_i \in [0,1]^{1 \times N}$ indicates the similarity between its feature and that of N existing reference instances. we then determine its pseudo class label as follows:
\begin{equation}
    \begin{aligned}
        &c_n, v = argmax_{C_I}(S_i) \\
        &\widetilde{Y}_{\hat{p_i}c_n} = \begin{cases}
        v*\eta, \quad &if\ v > T_{sim} \\
        0, \quad &otherwise
        \end{cases}
    \end{aligned}
\end{equation}
where \jun{$c_n$ and $v$ denotes the most similar category and corresponding similarity, respectively.} $\eta$ and $T_{sim}$ are scale factor and similarity threshold, respectively. We then obtain a new target category heatmap $\bar{Y} = Y + \widetilde{Y}$. Following~\cite{zhou2019objects}, we compute the classification loss as follows:
\begin{equation}
     L_{SPLG} = -\cfrac{1}{N} \sum_{yxc} \begin{cases}
    (1-\hat{Y}_{yxc})^{\gamma}\log(\hat{Y}_{yxc}), \bar{Y}_{yxc}=1 \\
        \begin{aligned}
        &(1-\bar{Y}_{yxc})^{\alpha}(\hat{Y}_{yxc})^{\gamma} \\
        &\log(1-\hat{Y}_{yxc})
        \end{aligned}, otherwise
    \end{cases}
    \label{eq:loss_pcl}
\end{equation}

\subsection{Pixel-level Group Contrastive Learning}
\label{sec:ContrastiveLearning}
%Recently, Contrastive Learning achieves great success in self-supervised task. To mitigate the reliance on pseudo-labels and boost the tolerance to false labels, \cite{wang2021self} proposes a Pseudo Group Contrast mechanism(PGC). Different from general Contrastive Learning which only has one positive pair for each image, it maintains a queue for each category and each image feature is pushed in the queue of class c when it has ground truth class label c or its pseudo class label is class c. However these methods are image-level Contrastive Learning for pretraining and need to maintain the feature queues.
The SPLG recalls several latent instances based on feature similarity, but the inaccurate pseudo labels inevitably raise the error amplification issue which is led by the cross-entropy loss~\cite{wang2021self}.
To overcome the drawbacks of class discrimination for self-training, we propose the Pixel-level Group Contrastive Learning (PGCL) to boost the tolerance to false pseudo labels by focusing on exploring the intrinsic structure of data.

Different from standard contrastive learning which involves just a positive key in each contrast \jun{at image-level}, PGCL introduces a group of positive keys with the same pseudo-class to contrast with all negative keys from other pseudo classes following~\cite{wang2021self}. 
% \jun{Note that the contrastive learning exists in each image.}
Given the class prediction $\hat{Y} \in \mathbb{R}^{\hat{H} \times \hat{W} \times C}$ \jun{for an image}, we first select top-$m$ instances (pixels) $\hat{P_p} = \{\hat{p_1}, \hat{p_2}, \cdots, \hat{p_m}\}$ as latent positive samples and
maintain a mask $M \in \{0,1\}^{N \times m}$ where $M_{ij}$ indicates whether the selected sample $\hat{p_j}$ belongs to the category $c_i$ according to its self-predicted label. We then gather corresponding feature vectors as encoded positive keys $f_k \in \mathbb{R}^{m \times D}$.
%gather corresponding feature vectors $\hat{F_{K}} \in \mathbb{R}^{m \times D}$. 
For each labeled reference instance of class $c_i$, the encoded query $f_q^{c_i}$ is obtained by extracting its central feature and its according primary positive key is encoded as $f_{k0}^{c_i}$ using Eq.~(\ref{eq:gfeat}), which is augmented by weighted summing neighbor pixels following Gaussian distribution. Formally, the overall objective of the PGCL is summarized as:
\begin{equation}
    \begin{aligned}
     L_{pgcl}=&-\cfrac{1}{m}\sum_{i=1}^{N}\sum_{j=1}^{m} M_{ij}\log \cfrac{exp(f_q^{c_i}\cdot f_k^j /\tau)}{Z_i} \\
     & - \cfrac{1}{N}\sum_{i=1}^{N}\log \cfrac{exp(f_q^{c_i}\cdot f_{k0}^{c_i}/\tau)}{Z_i} \\
     Z_i=& \sum_{j=1}^{m}exp(f_q^{c_i}\cdot f_k^j /\tau) + \sum_{z=1}^{N}exp(f_q^{c_i}\cdot f_{k0}^{c_z} /\tau) \\
    \end{aligned}
    \label{eq:loss_pgcl}
\end{equation}
where $\tau$ denotes the temperature scaling. 
Obviously, PGCL maximizes the similarity between the query $f_q^{c_i}$ and a corresponding group of positive keys $\{ f_{k0}^{c_i}, f_{k}^{j}|M_{ij}=1 \}$ for each category $c_i$. 
% In Eq.~(\ref{eq:loss_pgcl}), a softmax function is firstly applied to the similarity between $f_k$ and $f_q^{c_i}$ to generate the probability distribution. 
Positive keys $f_{k0}^{c_i}$ and $\{f_k^j|M_{ij}=1 \}$ with self-predicted class label $c_i$ will compete with each others. 
Although there are some false predictions in the selected group, those false instances will be defeated in the above competition, because their encoded features tend to be less similar to the query compared with that of true ones. 
Therefore, the model will be updated mainly by the gradients of true self-predicted labels and tend to avoid being misled by false self-predicted labels. 
Due to the strong tolerance to false self-predicted labels, PGCL effectively minimizes the distance between unlabeled instances and reference instance of same class and therefore improves the predicted scores for unlabeled instances. 

\subsection{Application to Other Detectors}
\label{sec:frcnn_fcos}
\begin{figure}
    \begin{center}
    \includegraphics[width=\linewidth]{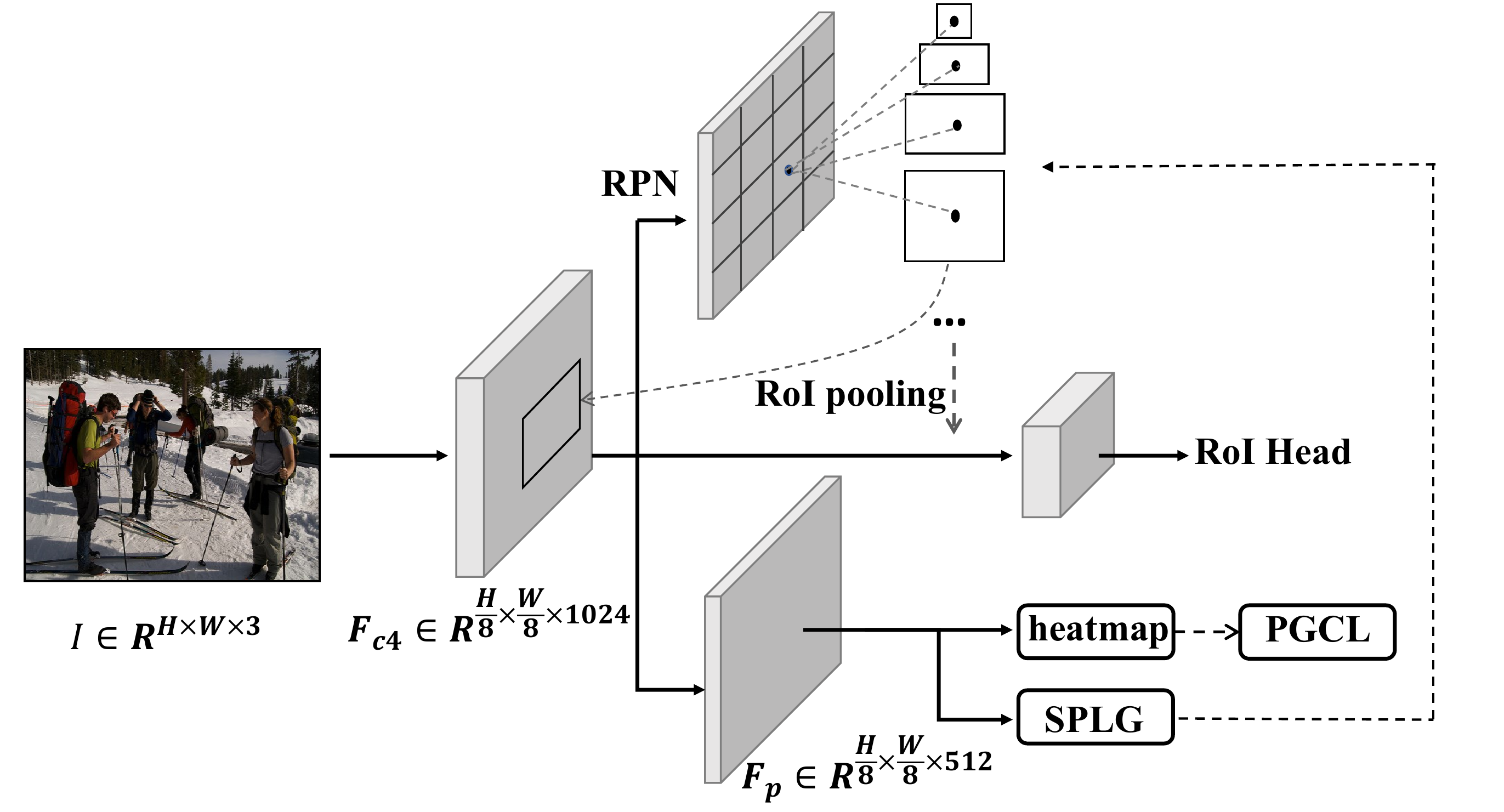}
    \end{center}
    \caption{Illustration of Faster-RCNN equipped with DMiner.}
    \label{fig:frcnn}
\end{figure}
In this section, we introduce how to apply the DMiner to other detectors, \eg two-stage anchor-based Faster-RCNN~\cite{ren2015faster} and muli-scale anchor-free FCOS~\cite{tian2019fcos}, on SIOD task.
For Faster-RCNN, we only apply the DMiner to the Region Proposal Network (RPN), because the classifier branch of RoI (Region of Interest) head is region-based while the DMiner is constructed on pixel-level features.
% For Faster-RCNN, we only apply the DMiner to the Region Proposal Network (RPN), because the classifier branch of RoI (Region of Interest) head from Faster-RCNN is region-based while the DMiner is constructed on pixel-level feature.
Since the RPN is class-agnostic, we cannot obtain specific categories for each anchor and the discrimination ability of the feature $F_{c4}$ constrained by RPN is limited, which leads to many inaccurate pseudo labels.
% In order to apply the DMiner\nothing{and enhance the discrimination ability of features}, 
We therefore introduce a new class-wise classifier branch in parallel with the RPN and apply the DMiner to the new branch, as shown in the Fig.~\ref{fig:frcnn}.
To assign satisfactory pseudo labels for each anchor, we utilize the average pooling operation on the initial pseudo labels generated by the feature map $F_p$ with different kernel sizes (\eg 1, 3, 5, 7, 9) according to the sizes of predefined anchors (\eg 32, 64, 128, 256, 512) to automatically obtain corresponding pseudo labels for each anchor. 

For FCOS, we directly apply the DMiner to each feature level of FPN~\cite{lin2017feature} structure in a similar way.
% since its prediction at each level is very similar to the CenterNet, DMiner can be directly apply to each feature utilized by FCOS independently. 
Concretely, the DMiner in each feature level shares hyperparamerters except to the number $m$ of selected positive samples in PGCL module, because the discrimination ability of feature reduces as the receptive filed being larger.
Especially, the features from P6, P7 level have limited discrimination due to the large overlap between receptive fields of each pixel-level feature.
In the practice, we only apply the DMiner at the first three level features and set the $m$ to $[96, 64, 32]$.
\section{Experiments}
\subsection{Datasets and Evaluation Protocol}
\noindent \textbf{Datasets.} According to the definition of SIOD, we first construct a new dataset called Keep1-COCO2017-Train via randomly preserving one annotation for each existing category in each image from the training set of COCO2017~\cite{lin2014microsoft}. In this way, it reduces about 60\% instance annotations. We keep the validation set COCO2017-Val the same as usual for comparison with fully supervised object detection.

\noindent \textbf{Evaluation protocol.} 
%\subsection{Score-aware Detection Evaluation Protocol}
%\label{sec:protocol}
% Since official COCO evaluation protocol cannot distinguish the ability of scoring between two different detectors for SIOD task (refer to supplemental materials for more details), 
Since the official COCO evaluation protocol cannot distinguish the difference of detection results with different scoring distributions (refer to supplemental materials for more details), 
we propose a new Score-aware Detection Evaluation Protocol for measuring such ability. Given a ground truth bounding box $g$ of class $c$ and a predicted bounding box $d$ of class $c$ with score $s_d$, we add a score constraint to official COCO matching rule and determine the match results as follows:
\begin{equation}
    M_{t_{iou}}^{t_{s}}(g,d,s_d) = \begin{cases}
        1, IoU(g,d) > t_{iou}\ and \ s_d > t_{s} \\
        0, \quad otherwise
    \end{cases}
\end{equation}
where $t_s \in \{0.,0.1,0.2,0.3,0.4,0.5,0.6,0.7,0.8,0.9\}$ is the score threshold and $t_{iou}$ is the IoU threshold. 
For brevity, we denote $AP@{\mathbb{S}_i}$ as the average precision over all IOU thresholds for score threshold $ t_s = i / 10$.
We additionally summarize a more comprehensive metric as follows:
\begin{equation}
    AP@\mathbb{S} = \frac{1}{10} \sum_{i=0}^{9} AP@\mathbb{S}_i 
\end{equation}
\begin{table*}[ht]
    \newcommand{\tabincell}[2]{\begin{tabular}{@{}#1@{}}#2\end{tabular}} % for newline
    \begin{center}
    \begin{tabular}{l|l|c|c|c|c|c|c}
        % \hline 
        \toprule[1pt]
        \multirow{2}{*}{Detector}& \multirow{2}{*}{Task} & \multicolumn{6}{c}{$AP(\%)$}  \\
        \cline{3-8} 
        & &  $@\mathbb{S}$ &  $@\mathbb{S}_0$  &  $@\mathbb{S}_3$  &  $@\mathbb{S}_5$  &  $@\mathbb{S}_7$   &  $@\mathbb{S}_9$  \\
        \hline 
        \multirow{3}{*}{\tabincell{c}{CenterNet-\\Res18~\cite{zhou2019objects}}} & FSOD & 17.3 & 28.1 & 24.0 & 17.1 & 8.8 & 1.5 \\ 
        & SIOD (base) & 13.9 & 25.1 & 18.5 & 12.3 & 6.1 & 1.4 \\ 
        & SIOD (DMiner) & 16.8 (+2.9) & 26.6 (+1.5) & 22.4 (+3.9) & 17.1 (+4.8) & 9.4 (+3.3) & 2.1 (+0.7) \\ 
        \hline 
        \multirow{3}{*}{\tabincell{c}{CenterNet-\\Res101~\cite{zhou2019objects}}} & FSOD & 22.6 & 34.2 & 30.3 & 23.6 & 13.6 & 3.1 \\ 
        & SIOD (base) & 15.1 & 27.8 & 20.9 & 13.3 & 6.1 & 1.1 \\ 
        & SIOD (DMiner) & 19.7 (+4.6) & 29.8 (+2.0) & 26 (+5.1) & 20.5 (+7.2) & 12.2 (+6.1) & 2.9 (+1.8) \\
        \hline
        \multirow{3}{*}{\tabincell{c}{Faster-RCNN- \\ Res50-C4~\cite{ren2015faster}}} & FSOD & 32.8 & 35.9 & 34.7 & 33.2 & 31.2 & 26.1 \\
        & SIOD (base) & 27.0 & 31.6 & 29.4 & 27.3 & 24.6 & 18.9 \\ 
        & SIOD (DMiner) & 29.2 (+2.2) & 31.9 (+0.3) & 30.6 (+1.2) & 29.5 (+2.2) & 27.8 (+3.2) & 23.9 (+5.0) \\ 
        \hline
        \multirow{3}{*}{\tabincell{c}{FCOS-Res50-\\FPN~\cite{tian2019fcos}}} & FSOD & 27.1 & 38.6 & 38.3 & 33.5 & 16.0 & 0.1 \\
        & SIOD (base) & 22.0 & 33.2 & 32.1 & 25.6 & 11.3 & 0 \\ 
        & SIOD (DMiner) & 23.6 (+1.6) & 33.9 (+0.7) & 33.3 (+1.2) & 28.6 (+3.0) & 14.1 (+2.8) & 0 \\
        % \hline
        \bottomrule[1pt]
    \end{tabular}
    \end{center}
    % \caption{Performance comparison across different detectors on COCO2017-Val for single instance annotated object detection (SIOD) and fully supervised object detection (FSOD) task. SIOD (base) denotes that we directly apply according detector to SIOD task and SIOD (DMiner) denotes that the detector is equipped with DMiner.}
    \caption{Detection results on COCO2017-Val for single instance annotated object detection (SIOD) and fully supervised object detection (FSOD) task. SIOD (base) denotes that we directly apply according detector to SIOD task and SIOD (DMiner) denotes that the detector is equipped with DMiner.}
    \label{tab:main}
\end{table*}
In this paper, we evaluate the detector with the proposed score-aware detection evaluation protocol and provide a more comprehensive comparison. We report the AP with different scores constraint $AP@\mathbb{S}_i$ (\eg $AP@\mathbb{S}_1$ for $t_s = 0.1$). 
Note that the $AP@\mathbb{S}_0$ is completely the same as the official COCO evaluation protocol. 

\noindent \textbf{Implementation details.} In this work, our experiments are mainly conducted with the CenterNet framework~\cite{zhou2019objects}. For CenterNet, we train on an input resolution of 512 $\times$ 512. This yields an output resolution of 128 $\times$ 128. Adam optimizer is utilized to optimize the network parameters. For CenterNet-Res18, we train with a batch-size of 114 (on 4 GPUs) and initial learning rate is 5e-4 for 140 epochs. For CenterNet-Res101, we train with a batch-size of 96 (on 8 GPUs) and initial learning rate is 3.75e-4 for 140 epochs. Both of them decay the learning rate with factor 0.1 at 90 and 120 epochs. As for Faster-RCNN-Res50-C4, we conduct the experiments with the detectron2 framework~\cite{wu2019detectron2}. FCOS-Res50-FPN is implemented with the official code of \cite{tian2019fcos}. All experiments are conducted with the environment that Tesla V100, Pytorch 1.7.0 and CUDA 10.1. As for hyper-parameters, $\eta$ and $T_{sim}$ are set to 1.0 and 0.6 by default, respectively. And $\lambda_{pgcl}, \lambda_{off}, \lambda_{size}$ are set to 0.1, 1.0, 0.1, respectively.  
\subsection{Main Results}
We first examine the effect on CenterNet-Res18 from FSOD to SIOD with the proposed score-aware detection evaluation protocol. Although the annotations have been reduced about 60\%, $AP@\mathbb{S}_0$  just decreases by $3.0$, which shows that the detector still has competitive localization ability under SIOD setup.
%We hypothesize that a detector can learn to locate well an object with moderate instance-level annotations and excessive annotations of each category will bring mild gain of performance.
However, the performance gap increases obviously (-5.5) on $AP@\mathbb{S}_3$, which explains the phenomenon in Fig.~\ref{fig:vis_res18_comp}(b). 
Compared $AP@\mathbb{S}_3$ with $AP@\mathbb{S}_0$, large gap indicates that the scores of a large number of detected objects are between $0$ and $0.3$, which are determined as false recall when the score threshold is set to $0.3$. 
A main reason is that most of unlabeled instances are treated as background during the training. 
\nothing{It obviously leads to a contradictory learning for a pair of instances of same class, namely, the labeled instance is treated as foreground and the other is treated as background.} 
To refine the incorrect supervision, the proposed SPLG mines the latent positive instances based on feature similarity.
In addition, the PGCL adopts the contrastive learning to boost the tolerance to false pseudo labels.
Equipped with two modules, CenterNet-Res18-SIOD (DMiner) improves the detection performances across different score thresholds consistently as shown in Table~\ref{tab:main}. 
Compared with SIOD (base), our method improves by 1.5 and 3.9 at $AP@\mathbb{S}_0$ and $AP@\mathbb{S}_3$, respectively. 
The former indicates that our method recalls more instances while the latter indicates that our method raises the scores of those low-quality detection (objects with scores less than $0.3$) up to $0.3$. 
Note that the gap between SIOD and FSOD is narrowed to $0.5$ on the  more comprehensive metric $AP@\mathbb{S}$. 
Other than verifying the effectiveness of DMiner with the small network (Res18), we further validate our method with a large backbone (Res101). 
As shown in Table~\ref{tab:main} (CenterNet-Res101), large improvements are obtained consistently. 
To validate the effectiveness on various detection frameworks, we conduct experiments with detector Faster-RCNN and FCOS. Both of them are confronted with incorrect background supervision and have large performance degradation. After equipped with the proposed DMiner, the performances are improved across different score thresholds.

From Table~\ref{tab:main}, we also observe that as the backbone network gets larger (from Res18 to Res101), the performance gaps between detectors on FSOD and SIOD (base) settings increase.
In our opinion, as the model gets larger, more labeled data is needed to learn optimal parameters. 
% Therefore, compare with Res18, large reduction of annotation affects Res101 more. 
Considering the computation cost, the proposed method only introduces extra cost in the training phase. 
Specifically, it requires about 1.4$\times$ time for learning and additional 3$\times$ memory compared with the baseline method.
% Note that the performance of CenterNet-Res101 drops more than that of CenterNet-Res18 from FSOD to SIOD(base). In our opinion, as the model gets larger, more labeled data is needed to learn stable parameters. Therefore, compare with Res18, large reduction of annotation affects Res101 more. Additionally, the proposed method only introduces extra cost in the training phase. Specifically, it just cost 1.4x time and requires 3x memory compared with the baseline method.

\begin{table}[t]
    \begin{center}
    \begin{tabular}{l|c|c|c|c|c}
        \toprule[1pt]
        \multirow{2}{*}{Method}& \multirow{2}{*}{Type} & \multicolumn{4}{c}{$AP(\%$)} \\
        \cline{3-6}
        &  & $@\mathbb{S}$ & $@\mathbb{S}_0$ & $@\mathbb{S}_3$ & $@\mathbb{S}_5$  \\
        \hline 
        Base & SSOD & 14.4 & 23.4 & 19.5 & 14.3  \\
        Base & SAOD & 14.0 & 25.0 & 18.6 & 12.7  \\
        Base & SIOD & 13.9 & 25.1 & 18.5 & 12.3  \\
        \hline
        CSD~\cite{jeong2019consistency} & SSOD & 15.1 & 24.0 & 20.3 & 15.1  \\
        TS~\cite{sohn2020simple} & SSOD & 15.8 & 25.2 & 21.4 & 15.8 \\
        Comining~\cite{wang2020co} & SAOD & 14.8 & 25.0 & 18.9 & 13.9 \\
        DMiner & SIOD & 16.8 & 26.6 & 22.4 & 17.1 \\
        \bottomrule[1pt]
    \end{tabular}
    \end{center}
    \caption{We implement different methods from semi-supervised object detection(SSOD) and sparsely annotated object detection(SAOD) with CenterNet-Res18 and evaluate them on COCO2017-Val for fair comparison.}
    \label{tab:method_comp}
\end{table}
% \begin{table}[h]
%     \begin{center}
%     \begin{tabular}{l|c|c|c|c|c|c|c}
%         \toprule
%         \multirow{2}{*}{Method}& \multirow{2}{*}{Type} & \multicolumn{6}{c}{AP} \\
%         \cline{3-8}
%         &  & s & s0 & s3 & s5 & s7 & s9 \\
%         \hline 
%         \hline
%         Base & SSOD & 14.4 & 23.4 & 19.5 & 14.3 & 7.6 & 1.6 \\
%         Base & SAOD & 14.0 & 25.0 & 18.6 & 12.7 & 6.5 & 1.7 \\
%         Base & SIOD & 13.9 & 25.1 & 18.5 & 12.3 & 6.1 & 1.4 \\
%          \hline
%         CSD & SSOD & 15.1 & 24 & 20.3 & 15.1 & 8.5 & 1.8 \\
%         TS & SSOD & 15.8 & 25.2 & 21.4 & 15.8 & 8.5 & 1.8 \\
%         Comining & SSOD & & & & & \\
%         Ours & SIOD & 16.8 & 26.6 & 22.5 & 17.2 & 8.3 & 2.0 \\
%         \bottomrule
%     \end{tabular}
%     \end{center}
%     \caption{Caption}
%     \label{tab:my_label}
% \end{table}
\subsection{Comparison with Other Methods} 
To fairly compare with other methods proposed for SSOD (\eg CSD~\cite{jeong2019consistency}) or SAOD (Comining~\cite{wang2020co}), we implement them with CenterNet-Res18. Specifically, we first construct a new training set for SSOD methods via randomly preserving a number of completely annotated images and erasing all annotations for the rest images from COCO2017-Train, which has equivalent instance annotations to Keep1-COCO2017-Train. As shown in Table~\ref{tab:method_comp}, Base denotes directly training the detector on according training set. Since SIOD reduces the inter-image discrepancy to intra-image discrepancy, %and provides relatively balanced annotations, 
the baseline of SIOD is obviously superior to SSOD at $AP@\mathbb{S}_0$,
%the baseline at $AP@\mathbb{S}_0$ of SIOD is superior to that of SSOD,
which demonstrates that the proposed annotated manner retrains richer information. 
%Since SIOD reduces the inter-image discrepancy to intra-image discrepancy and provides rich prior knowledge(\eg reference instance), DMiner which is designed according such characteristic, outperforms CDS and TS~\cite{sohn2020simple} (Teacher-Student method)
Furthermore, the proposed method outperforms CSD and TS (Teacher-Student)
%wchich the teacher is trained with the labeled data in advance and student imitate the prediction of teacher on unlabeled data.
by a large margin with same number of instance annotations. As for Comining, it is very sensitive to score threshold for selecting pseudo positive samples and tends to fall into collapse. Therefore, the improvement brought by Comining is very limited. 
\begin{table}[t]
    \begin{center}
    \begin{tabular}{l|c|c|c|c|c|c}
        \toprule[1pt] 
         \multirow{2}{*}{Backbone} & \multicolumn{2}{c|}{Module} & \multicolumn{4}{c}{$AP(\%$)}  \\
         \cline{2-7} 
         & SPLG & PGCL & $@\mathbb{S}$ & $@\mathbb{S}_0$ & $@\mathbb{S}_3$ & $@\mathbb{S}_5$\\
         \hline 
         \multirow{4}{*}{Res18}& & & 13.9 & 25.1 & 18.5 & 12.3 \\ 
         & \checkmark & & 15.8 & 26.4 & 21.5 & 15.7 \\ 
         &  & \checkmark & 16.2 & 25.8 & 22.3 & 16.6  \\
         & \checkmark & \checkmark & 16.8 & 26.6 & 22.4 & 17.1 \\
         \hline 
         \multirow{4}{*}{Res101}& & & 15.1 & 27.8 & 20.9 & 13.3  \\
         & \checkmark & & 18.5 & 30 & 25.2 & 18.6 \\
         &  & \checkmark & 18.9 & 29.2 & 25.8 & 20.1 \\
         & \checkmark & \checkmark & 19.7 & 29.8 & 26 & 20.5  \\
         \bottomrule[1pt] 
    \end{tabular}
    \end{center}
    \caption{Effectiveness of SPLG and PGCL for SIOD task on COCO2017-Val.}
    \label{tab:ab_pcl_gcl}
\end{table}
% \begin{table*}[ht]
%     \begin{center}
%     \begin{tabular}{l|c|c|c|c|c|c|c|c}
%         \toprule[1pt] 
%          \multirow{2}{*}{Detector} & \multicolumn{2}{c|}{Module} & \multicolumn{6}{c}{AP}  \\
%          \cline{2-9} 
%          & PCL & PGCL & s & s0 & s3 & s5 & s7 & s9 \\
%          \hline 
%          \multirow{4}{*}{CenterNet-Res18}& & & 13.9 & 25.1 & 18.5 & 12.3 & 6.1 & 1.4 \\ 
%          & \checkmark & & 15.7 & 26.2 & 21.5 & 15.7 & 7.5 & 1.2 \\ 
%          &  & \checkmark & 16.2 & 25.9 & 22.3 & 16.8 & 8.5 & 1.2 \\
%          & \checkmark & \checkmark & 16.8 & 26.6 & 22.5 & 17.2 & 8.3 & 2.0 \\
%          \hline 
%          \multirow{4}{*}{CenterNet-Res101}& & & 15.1 & 27.8 & 20.9 & 13.3 & 6.1 & 1.1 \\
%          & \checkmark & & 18.5 & 30 & 25.2 & 18.6 & 9.7 & 1.5 \\
%          &  & \checkmark & 18.9 & 29.2 & 25.8 & 20.1 & 10.6 & 1.6 \\
%          & \checkmark & \checkmark & 19.7 & 29.8 & 26 & 20.5 & 12.2 & 2.9 \\
%          \bottomrule[1pt] 
%     \end{tabular}
%     \end{center}
%     \caption{Effectiveness of PCL and PGCL for SIOD task on COCO2017-Val.}
%     \label{tab:ab_pcl_gcl}
% \end{table*}
\subsection{Ablations Experiments}
\paragraph{Effectiveness of SPLG and PGCL.} To validate the effectiveness of the proposed SPLG and PGCL, we equip the basic detector (CenterNet) with them for training respectively. As shown in Table \ref{tab:ab_pcl_gcl}, both SPLG and PGCL boost the performance independently. Specifically, SPLG tends to improve the $AP@\mathbb{S}_0$ while PGCL improves the $AP@\mathbb{S}$ greatly, which means that SPLG is beneficial to mining more instances and PGCL tends to improve the scores of latent instances. After integrating them, the detector achieves a higher performance.
\paragraph{Hyper-parameters Selection for SPLG.} Note that the accuracy of pseudo labels 
is very sensitive to the $T_{sim}$. A higher score threshold will lead to low recall rate of positive samples and a lower score threshold will lead to generating a large number of false pseudo class labels. As shown in Table~\ref{tab:ab_pcl}, although it achieves the best performance with $T_{sim}=0.5$, it actually almost determines all candidate positions as foregrounds, which is unreasonable. Knowing that PGCL module will maximize the similarity between positive pairs, we finally set the $T_{sim}$ to $0.6$ for better performance in combination with PGCL by default. As for the scaling factor $\eta$, large value tends to result in model collapse and no scaling is the best choice. 
\newcommand\visfigurewidth{0.195}
\begin{figure*}
\centering

    \includegraphics[width=\visfigurewidth\linewidth]{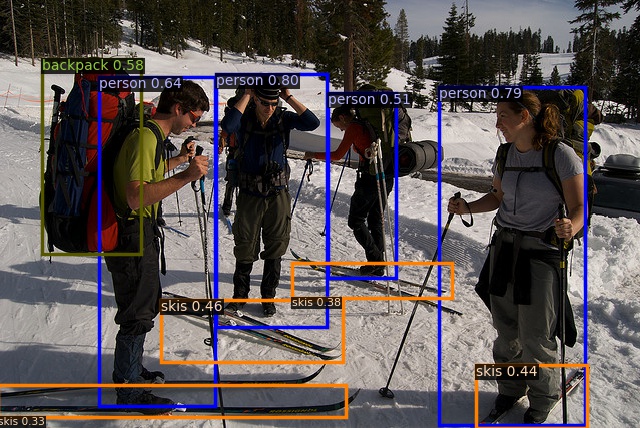}
    \includegraphics[width=\visfigurewidth\linewidth]{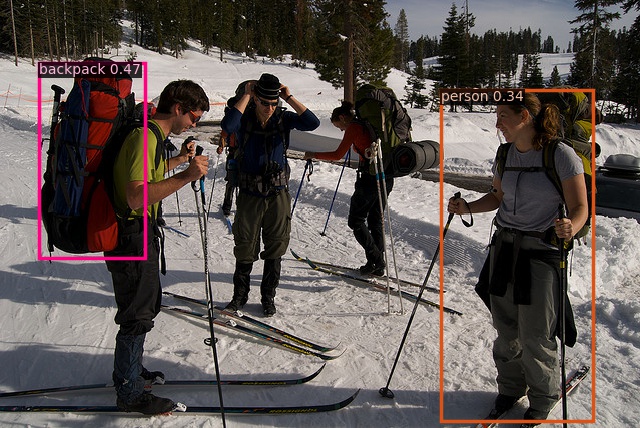}
    \includegraphics[width=\visfigurewidth\linewidth]{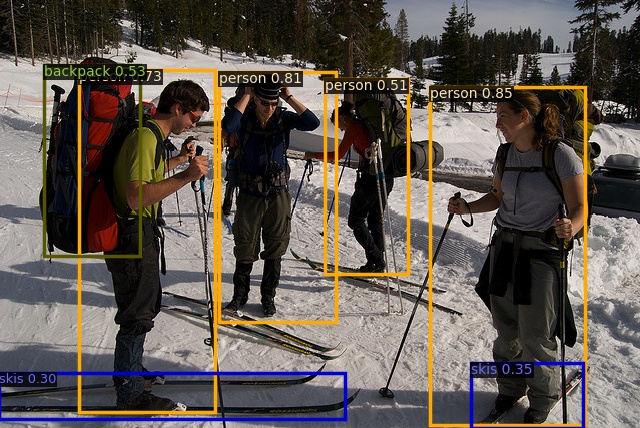}
    \includegraphics[width=\visfigurewidth\linewidth]{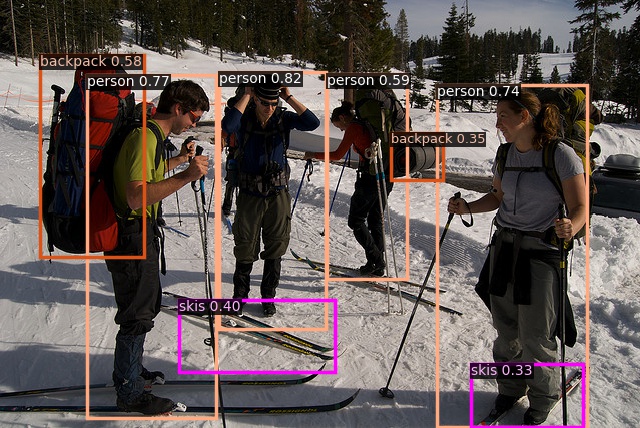}
    \includegraphics[width=\visfigurewidth\linewidth]{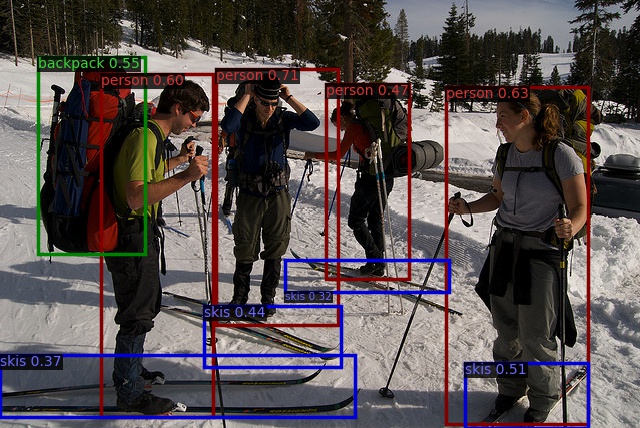} \\
     \includegraphics[width=\visfigurewidth\linewidth]{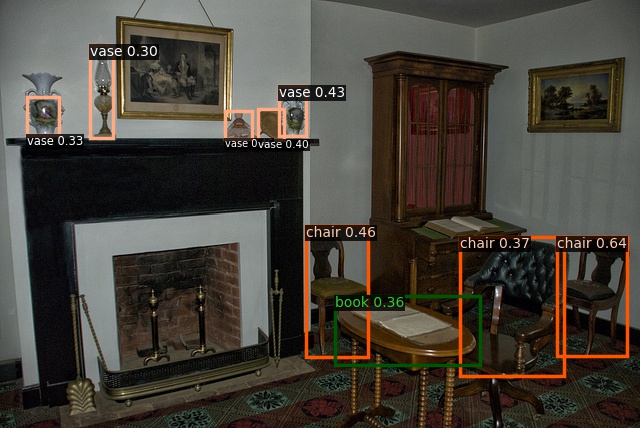}
    \includegraphics[width=\visfigurewidth\linewidth]{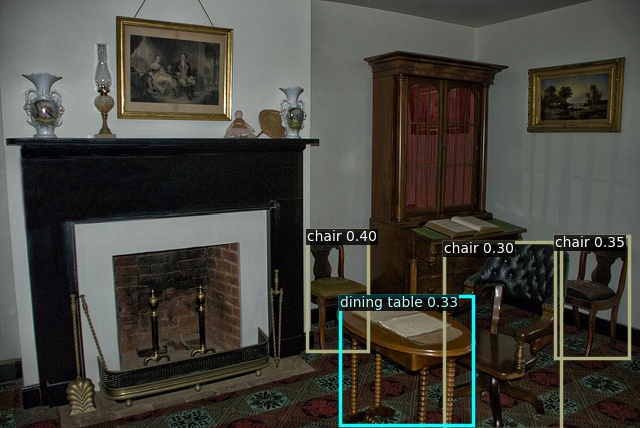}
    \includegraphics[width=\visfigurewidth\linewidth]{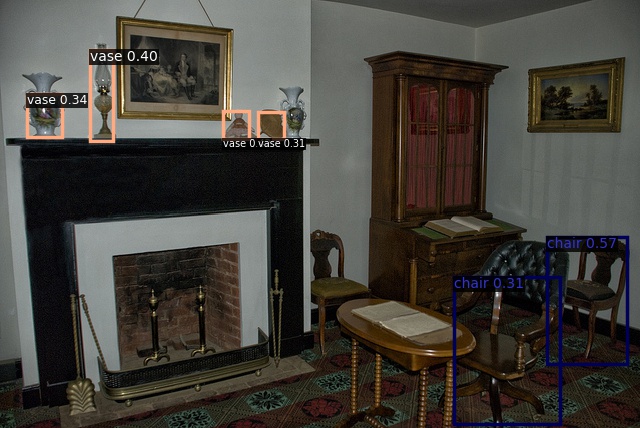}
    \includegraphics[width=\visfigurewidth\linewidth]{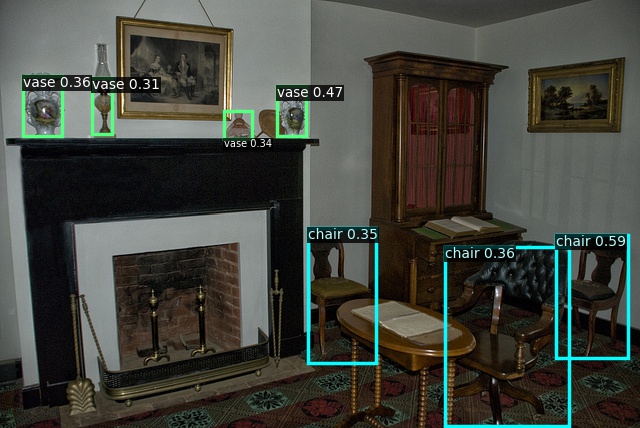}
    \includegraphics[width=\visfigurewidth\linewidth]{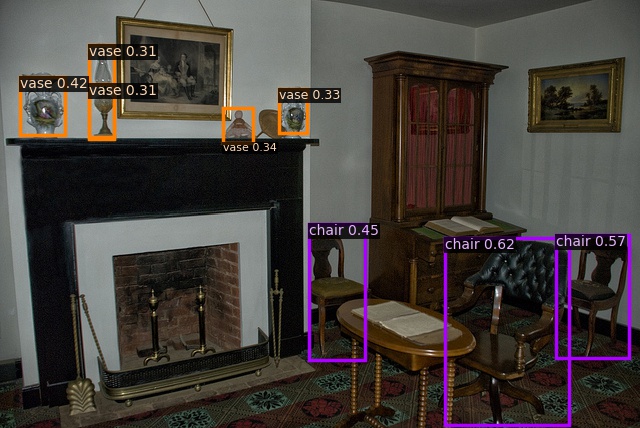} \\
     \includegraphics[width=\visfigurewidth\linewidth]{Images/vis_res/fsod_all_000000013546.jpg}
    \includegraphics[width=\visfigurewidth\linewidth]{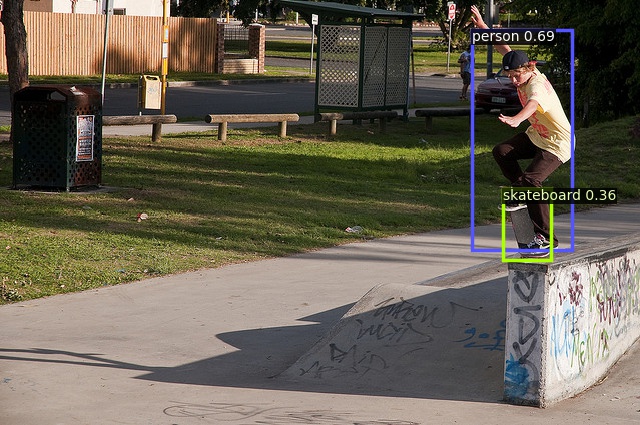}
    \includegraphics[width=\visfigurewidth\linewidth]{Images/vis_res/ssod_all_000000013546.jpg}
    \includegraphics[width=\visfigurewidth\linewidth]{Images/vis_res/ssod_ts_all_000000013546.jpg}
    \includegraphics[width=\visfigurewidth\linewidth]{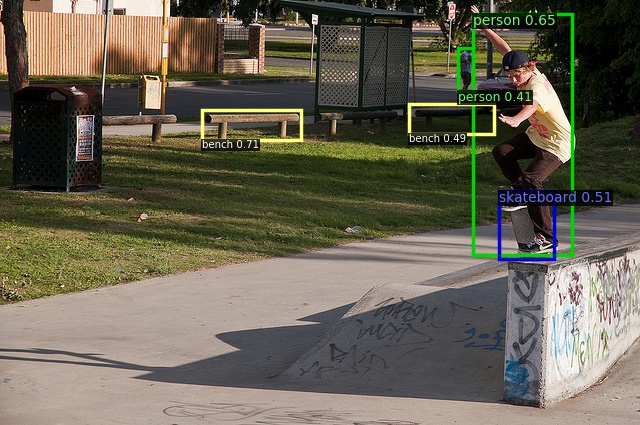} \\
    \begin{minipage}{\visfigurewidth\linewidth}
    \centering
    \figtext{(a) FSOD}
    \end{minipage}
    \begin{minipage}{\visfigurewidth\linewidth}
    \centering
    \figtext{(b) SIOD (base)}
    \end{minipage}
    \begin{minipage}{\visfigurewidth\linewidth}
    \centering
    \figtext{(c) SSOD-CSD}
    \end{minipage}
    \begin{minipage}{\visfigurewidth\linewidth}
    \centering
    \figtext{(d) SSOD-TS}
    \end{minipage}
    \begin{minipage}{\visfigurewidth\linewidth}
    \centering
    \figtext{(e) Ours}
    \end{minipage}
    \caption{Visualization of different methods with CenterNet-Res18 with score threshold 0.3, (a)FSOD, (b) SIOD(base), 
    (c) SSOD-CSD, (d) SSOD-TS, (e) Ours.}
    \label{fig:vis_res18_comp}
\end{figure*}
\begin{table}[t]
    \begin{center}
    \begin{tabular}{l|c|c|c|c|c}
        \toprule[1pt]
        $T_{sim}$ & 0.5 & 0.55 & 0.6 & 0.65 & 0.7  \\
        \hline
        $AP@\mathbb{S}$ & 16.2 & 16 & 15.8 & 15.5 & 15.4 \\
        $AP@\mathbb{S}_0$ & 26.8 & 26.5 & 26.4 & 26 & 26.1 \\
        \hline
        \hline
        $\eta$ & 0.8 & 0.9 & 1.0 & 1.1 & 1.2 \\
        \hline
        $AP@\mathbb{S}$ & 15.4 & 15.8 & 15.8 & 1.6 & 0 \\
        $AP@\mathbb{S}_0$ & 26.1 & 26.2 & 26.4 & 4.1 & 0 \\
        % warmup & 0 & 5 & 10 & 15 & 20 \\
        % \hline 
        % AP@s & 15.7 & 15.8 & 15.9 & 15.6 & 15.6 \\
        % AP@s0 & 26.2 & 26.3 & 26.7 & 26.1 & 26.2 \\
        \bottomrule[1pt] 
    \end{tabular}
    \end{center}
    \caption{Impact of different hyper-parameters in SPLG.}
    \label{tab:ab_pcl}
\end{table}
\begin{table}[t]
    \begin{center}
    \begin{tabular}{l|c|c|c|c|c}
        \toprule[1pt]
        $m$ & 64 & 96 & 128 & 160 & 192 \\
        \hline
        $AP@\mathbb{S}$ & 15.8 & 16.3 & 16.2 & 16.1 & 16 \\
        $AP@\mathbb{S}_0$ & 25.7 & 25.7 & 25.8 & 25.5 & 25.3 \\
        % \hline
        % \hline
        % neg-k & 0 & 32 & 64 & 96 & 128 \\
        % \hline 
        % AP@s & 16.2 & 16.3 & 16.1 & 16.3 & 16.3  \\
        % AP@s0 & 25.8 & 25.8 & 25.7 & 25.8 & 25.8 \\
        \hline
        \hline
        $\tau$ & 0.01 & 0.04 & 0.07 & 0.1 &  0.13 \\
        \hline
        $AP@\mathbb{S}$ & 0 & 16.1 & 16.2 & 16 & 15.7 \\
        $AP@\mathbb{S}_0$ & 0 & 25.8 & 25.8 & 25.6 & 25.2 \\
        \hline
        \hline 
        $\lambda_{pgcl}$ & 0.05 & 0.1 & 0.15 & 0.2 & 0.25 \\
        \hline 
        $AP@\mathbb{S}$ & 16.1 & 16.2 & 16.2 & 16.2 & 15.9 \\
        $AP@\mathbb{S}_0$ & 25.9 & 25.8 & 25.5 & 25.6 & 25.4 \\ 
        \bottomrule[1pt] 
    \end{tabular}
    \end{center}
    \caption{Impact of different hyper-parameters in PGCL.}
    \label{tab:ab_pgcl}
\end{table}
\paragraph{Hyper-parameters Selection for PGCL.} We first explore the effect of top-$m$ in PGCL. Since we determine the top-$m$ positions as foregrounds, many backgrounds will be recognized as foregrounds with large $m$ while few of foregrounds are recalled with small $m$. To better trade-off this contradiction, we explore different values from 64 to 192 and find that 128 is relatively suitable choice. As for the $\tau$ in Eq.~(\ref{eq:loss_pgcl}), the best performance is achieved when it is set to $0.07$, a value usually selected in Contrastive Learning by default. To better balance the effect of $L_{pgcl}$ and other losses, we try to opt different values for experiments. As shown in Table~\ref{tab:ab_pgcl}, the detector achieves satisfactory performance when $\lambda_{pgcl}$ is set to $0.1$.   
\paragraph{Advantage of SIOD Task.} Note that SIOD task is very similar to the SAOD task. We therefore conduct rigorous experiments for comparison. There are three different annotation sets (\eg easy, hard and extreme) for SAOD task. Specifically, the hard set keeps 50\% instance annotations from COCO2017-Train while only 40\% instance annotations are preserved in SIOD task. As shown in Table~\ref{tab:comining}, the method Comining further achieves better performance with the annotations from SIOD task compared with that of SAOD task, although less instance annotations are available. Such phenomenon suggests that the annotated manner of SIOD has great potential to achieve better performance with lower annotated cost. \nothing{We attribute the advantage of SIOD task to the annotated manner, which it annotates one instance for each category in each images all the time. }
\begin{table}[h]
    \begin{center}
    \begin{tabular}{l|c|c|c|c|c}
        \toprule[1pt]
        \multirow{2}{*}{Method} & \multirow{2}{*}{Type} & \multicolumn{4}{c}{$AP(\%)$} \\
        \cline{3-6}
        &  & $@\mathbb{S}$ & $@\mathbb{S}_0$  & $@\mathbb{S}_3$  & $@\mathbb{S}_5$   \\
        \hline
        Comining\cite{wang2020co} & SAOD & 19.4 & 31.6 & 26.8 & 16.7  \\
        Comining\cite{wang2020co} & SIOD & 23.6 & 32.4 & 28.5 & 23.6   \\
        \bottomrule[1pt]
    \end{tabular}
    \end{center}
    \caption{Experiments are conducted with RetinaNet-Res50 on two different annotation sets. Note that SAOD here adopts the \textbf{hard} sparse annotation sets \cite{wang2020co} for training.}
    \label{tab:comining}
\end{table}
% \begin{table}[h]
%     \begin{center}
%     \begin{tabular}{l|c|c|c|c|c|c|c}
%         \hline
%         \multirow{2}{*}{Method} & \multirow{2}{*}{Type} & \multicolumn{6}{c}{AP} \\
%         \cline{3-8}
%         &  & s & s0 & s3 & s5 & s7 & s9 \\
%         \hline
%         Comining & SAOD & 19.4 & 31.6 & 26.8 & 16.7 & 10.0 & 5.1 \\
%         Comining & SIOD & 23.6 & 32.4 & 28.5 & 23.6 & 17.9 & 9.9  \\
%         \hline 
%     \end{tabular}
%     \end{center}
%     \caption{Caption}
%     \label{tab:comining}
% \end{table}
\subsection{Visualization}
To clearly reveal the effectiveness of our method, we visualize the detection results of different methods under the proposed SIOD setup. 
As shown in Fig.~\ref{fig:vis_res18_comp}, column (a) shows the detection results of CenterNet-Res18 trained with fully annotated data, where most of instances are located and classified accurately.
However, the detection results of the detector directly trained for SIOD task are very unsatisfactory as shown in column (b). A main reason results in such phenomenon is that the incorrect background supervision confuses the detector during the training. Therefore, most of instances are detected with very low scores and are filtered when visualizing with $t_s=0.3$. Additionally, we also visualize two semi-supervised methods, column (c) and (d). From the first row, it is observed that they fail to locate some skis in the pictures. After armed with SPLG and PGCL of DMiner, most of salient instances are accurately detected with relatively higher scores compared with SIOD (base) as shown in column (e). Compared with DMiner, the locating of method CSD is relatively imperfect. As shown in bottom picture, the boy and the skateboard are detected partially by the method CSD.

\section{Conclusion}
% In this study, we investigate a new task called Single Instance annotated Object Detection(SIOD) to trade off the annotating cost and detection performance, which requires only one instance for each category.
% Compared to WSOD, SSOD, and SAOD, SIOD reduces the inter-task or inter-image discrepancies to intra-image discrepancy and provides more prior knowledge for exploiting unlabeled data with less annotation. 
In this study, we investigate a new task, termed Single Instance annotated Object Detection(SIOD).\nothing{, which requires only one instance for each category.
Compared to WSOD, SSOD and SAOD, SIOD reduces the inter-task or inter-image discrepancies to intra-image discrepancy and trade off the annotating cost and detection performance.}
Under the SIOD setup, we propose a simple yet effective framework, termed Dual-Mining (DMiner), to mine latent instances based on feature similarity by the proposed SPLG and further boost the tolerance to false pseudo labels by the proposed PGCL.  
The extensive experiments verify the superiority of the SIOD setup and the proposed DMiner effectively reduces the gap from the fully supervised object detection.
% Our experimental results show that both PCL and PGCL in DMiner effectively mine unlabeled instances for training and mitigate the incorrect supervision on unlabeled regions. 
% The performance gap compared with FSOD has been largely reduced while the cost of annotations is reduced largely. 
% To some extent, SIOD task is more likely applied to commercial application with such trade-off performance and the cost of annotations compared with WSOD. 
As the first and solid baseline under SIOD setup, DMiner provides a fresh insight to the challenging object detection task under imperfect data.

\noindent\textbf{Limitations.} Extensive experiments show that the SIOD task provides a promising way to trade off the annotation cost and detection accuracy, but the annotation cost of SIOD is still not to be ignored. It is worth to explore annotating only one instance in an image in the future work.  
Although we have witnessed the effectiveness of the proposed DMiner framework for SIOD task, the proposed approach remains challenging when applied to anchor-based architectures (\eg Faster RCNN). 
Since the SPLG and PGCL modules are constructed on pixel-level features, it is difficult to assign accurate pseudo labels for anchors that share the same feature representation. 
In the future, we will update the framework to better adapt to different architectures.
\section{Acknowledgements}
This work was supported partially by the NSFC(U21A20471,U1911401,U1811461), Guangdong NSF Project (No. 2020B1515120085, 2018B030312002), and the Key-Area Research and Development Program of Guangzhou (202007030004).

{\small
\bibliographystyle{ieee_fullname}
\bibliography{egbib}
}
\newcommand\fourcolwidth{0.245}
\begin{figure*}
    \centering
        \includegraphics[width=\fourcolwidth\linewidth]{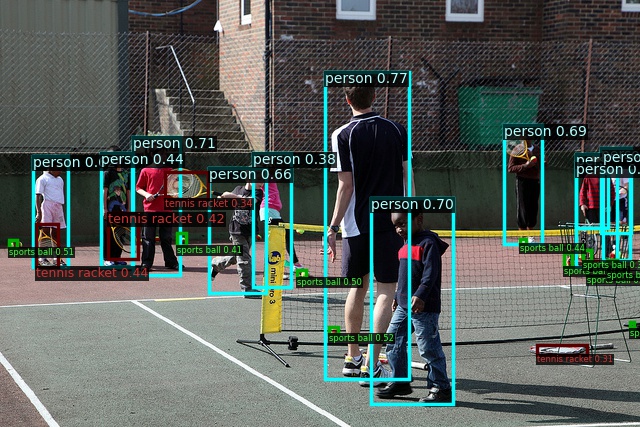}
        \includegraphics[width=\fourcolwidth\linewidth]{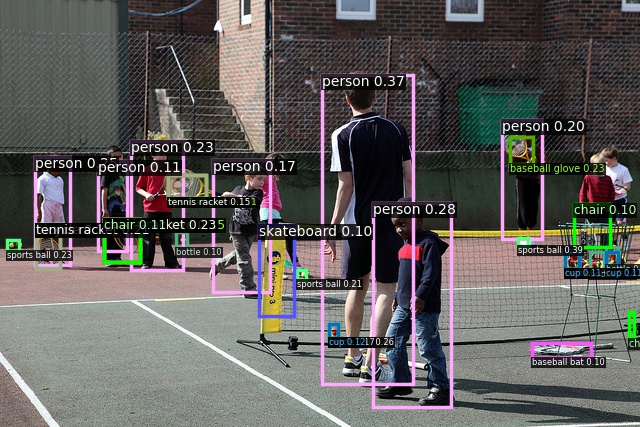}
        \includegraphics[width=\fourcolwidth\linewidth]{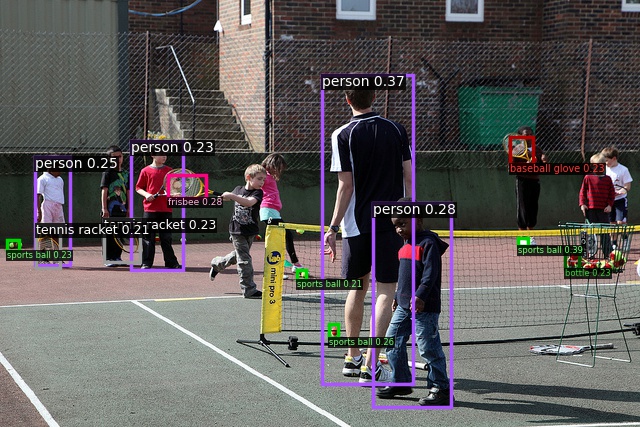}
        \includegraphics[width=\fourcolwidth\linewidth]{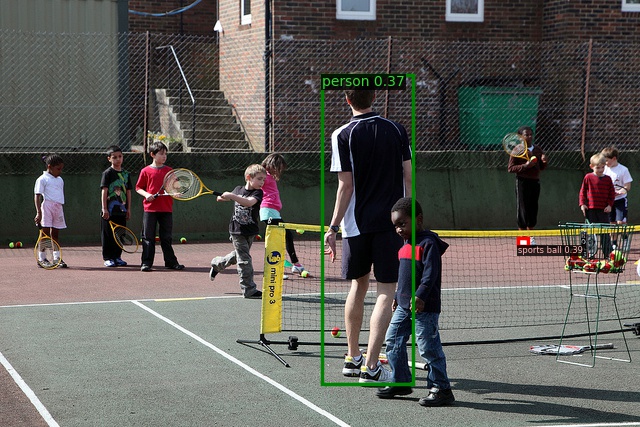} \\
        \includegraphics[width=\fourcolwidth\linewidth]{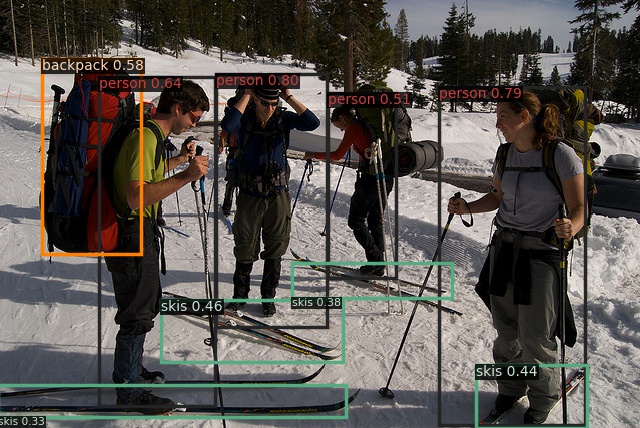}
        \includegraphics[width=\fourcolwidth\linewidth]{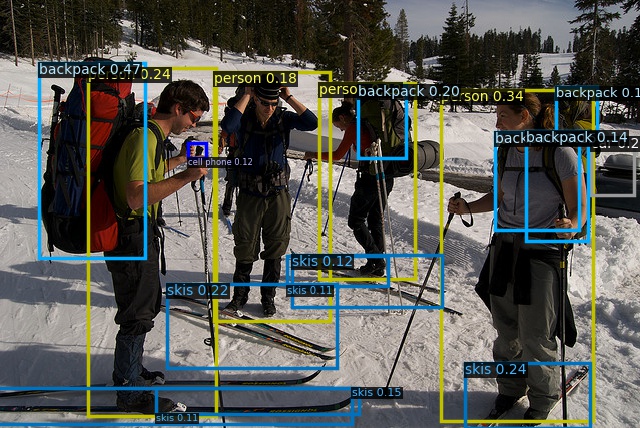}
        \includegraphics[width=\fourcolwidth\linewidth]{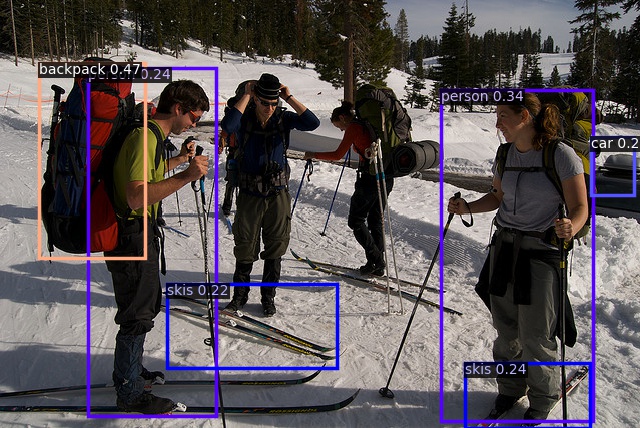}
        \includegraphics[width=\fourcolwidth\linewidth]{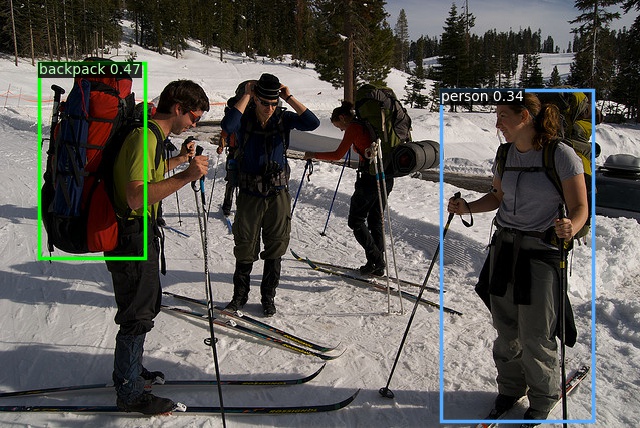} \\
        \includegraphics[width=\fourcolwidth\linewidth]{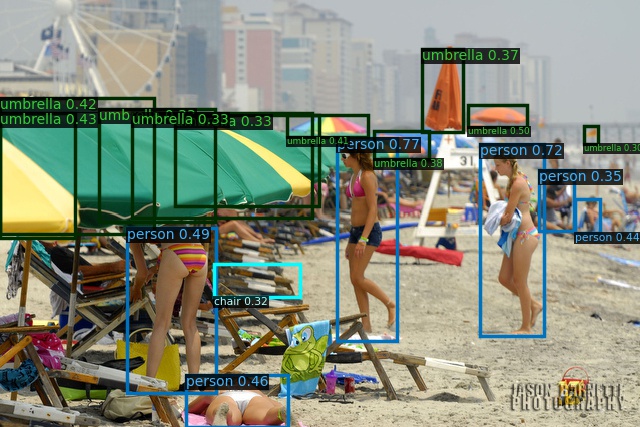}
        \includegraphics[width=\fourcolwidth\linewidth]{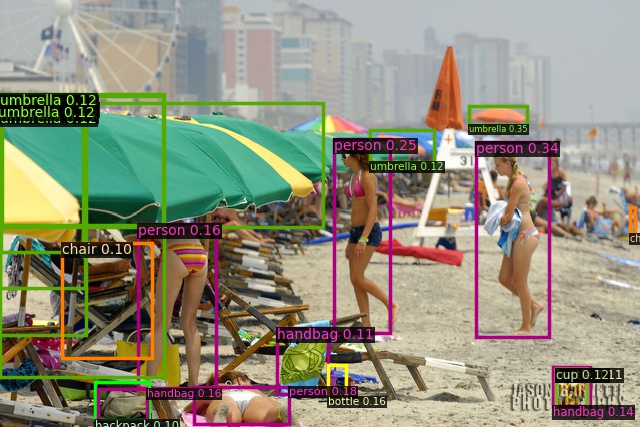}
        \includegraphics[width=\fourcolwidth\linewidth]{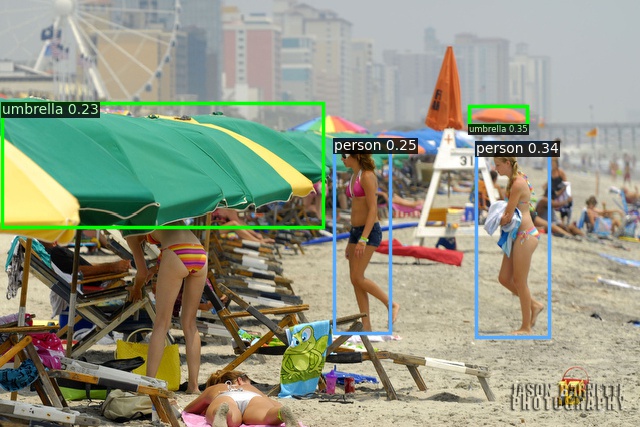}
        \includegraphics[width=\fourcolwidth\linewidth]{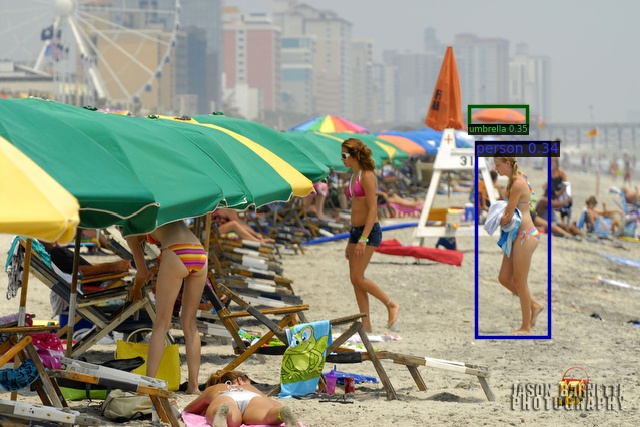} \\
        \begin{minipage}{\fourcolwidth\linewidth}
            \centering
            \figtext{(a) FSOD$@\mathbb{S}_3$}
        \end{minipage}
        \begin{minipage}{\fourcolwidth\linewidth}
            \centering
            \figtext{(b) SIOD(base)$@\mathbb{S}_1$}
        \end{minipage}
        \begin{minipage}{\fourcolwidth\linewidth}
            \centering
            \figtext{(c) SIOD(base)$@\mathbb{S}_2$}
        \end{minipage}
        \begin{minipage}{\fourcolwidth\linewidth}
            \centering
            \figtext{(d) SIOD(base)$@\mathbb{S}_3$}
        \end{minipage}
        \caption{Visualization of FSOD and SIOD(base) with CenterNet-Res18 across different score thresholds. Note that SIOD(base) denotes directly training the detector on SIOD task and $\mathbb{S}_i$ denotes the score threshold is $i/10$.}
        \label{fig:fsod_keep1}
\end{figure*}
\section{Supplementary Material}
% main manuscript 
\subsection{Meaning of Score-aware Detection Evaluation Protocol}
\begin{table}[h]
    \begin{center}
    \begin{tabular}{l|c|c}
        \toprule
        Detector &  Task & AP(\%) \\ 
        \hline
        \multirow{2}{*}{CenterNet-Res18} & FSOD & 28.1 \\
        & SIOD & 25.1 \\
        \hline
        \multirow{2}{*}{CenterNet-Res101} & FSOD & 34.2 \\
        & SIOD & 27.8 \\
        \bottomrule 
    \end{tabular}
    \end{center}
    \caption{The changes of AP from FSOD to SIOD task with CenterNet framework.
    The performance is evaluated on COCO2017-Val.}
    \label{tab:ap_drop}
\end{table}
In this section, we dive into analyzing the defect of COCO style evaluation protocol when it is applied to SIOD task. We first evaluate the performance of detector trained on FSOD and SIOD task, respectively. As shown in Table~\ref{tab:ap_drop}, it seems that the detector still performs well on SIOD task, although only 40\% instance annotations are preserved compared with FSOD task. Actually, the discriminative ability of two detectors (\eg CenterNet-Res18 trained on FSOD task or SIOD task) is still significantly different. We first visualize the detected bounding boxes with score threshold 0.3, as shown in Fig.~\ref{fig:fsod_keep1} column (a) and column (d). Few objects are detected when the detector is trained on SIOD task. As we decrease the score threshold, an increasing number of boxes are shown(\eg SIOD(base)$@\mathbb{S}_1$ and SIOD(base)$@\mathbb{S}_2$).
Obviously, SIOD(base) can achieve comparable performance with FSOD regardless of the score(confidence). Since official COCO evaluation protocol determines a true match without considering the predicted scores, a large number of detected bounding boxes with low scores are recalled (similar to Fig.~\ref{fig:fsod_keep1} SIOD(base)$@\mathbb{S}_1$ ).  In this way, it results in illusory advances on SIOD task. In order to distinguish the ability of scoring between two different detectors, we propose a Score-aware Detection Evaluation Protocol, which introduces a score constraint to the match rule of official COCO evaluation protocol. In this way, we can measure the performance of different detectors across different score thresholds. Undoubtedly, a perfect detector is expected to detect objects with high scores. The proposed evaluation protocol exactly is capable to measure such ability.  
\subsection{Visualization for SPLG and PGCL }
\begin{figure*}
    \centering
        \includegraphics[width=\fourcolwidth\linewidth]{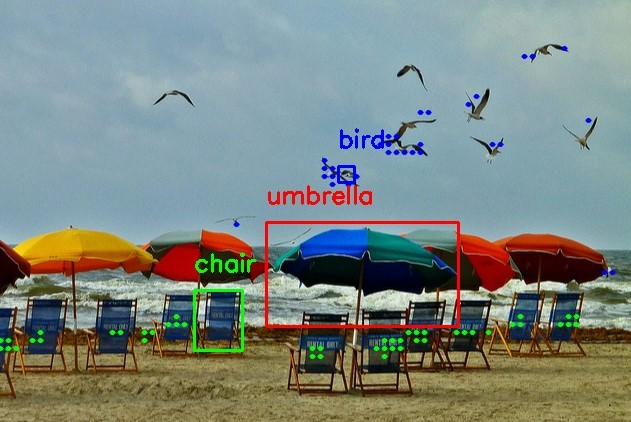}
        \includegraphics[width=\fourcolwidth\linewidth]{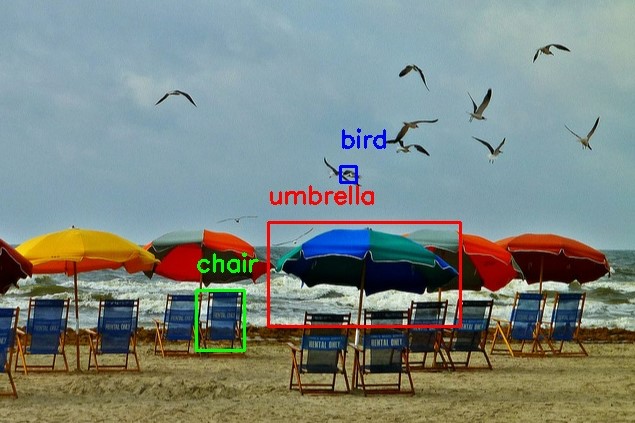}
        \includegraphics[width=\fourcolwidth\linewidth]{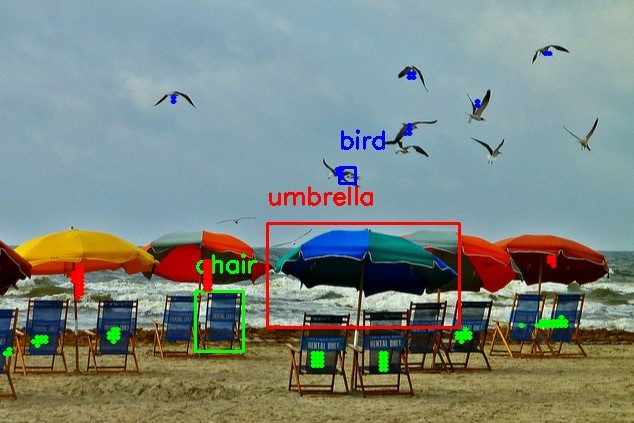}
        \includegraphics[width=\fourcolwidth\linewidth]{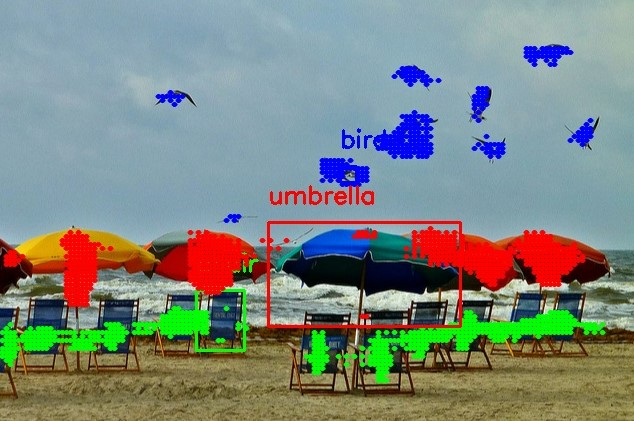} \\
        \includegraphics[width=\fourcolwidth\linewidth]{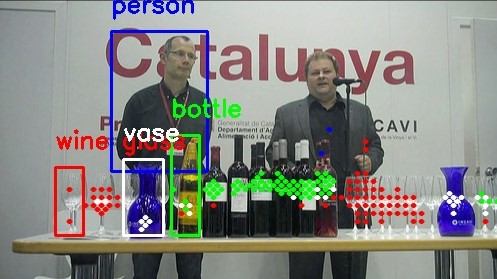}
        \includegraphics[width=\fourcolwidth\linewidth]{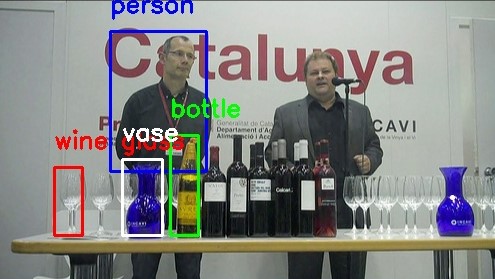}
        \includegraphics[width=\fourcolwidth\linewidth]{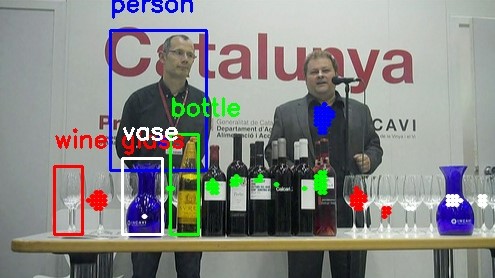}
        \includegraphics[width=\fourcolwidth\linewidth]{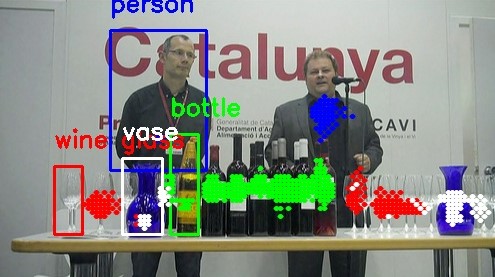} \\
        \includegraphics[width=\fourcolwidth\linewidth]{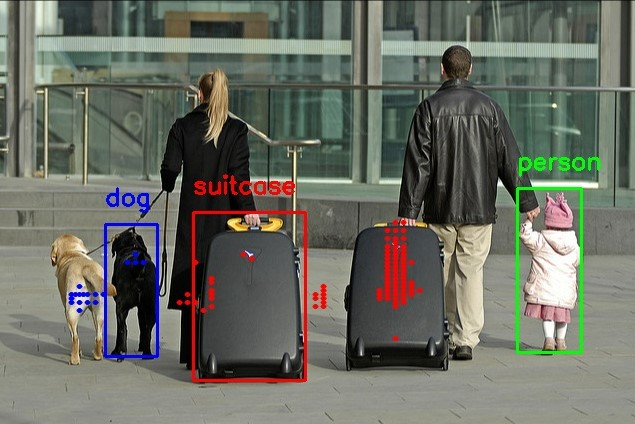}
        \includegraphics[width=\fourcolwidth\linewidth]{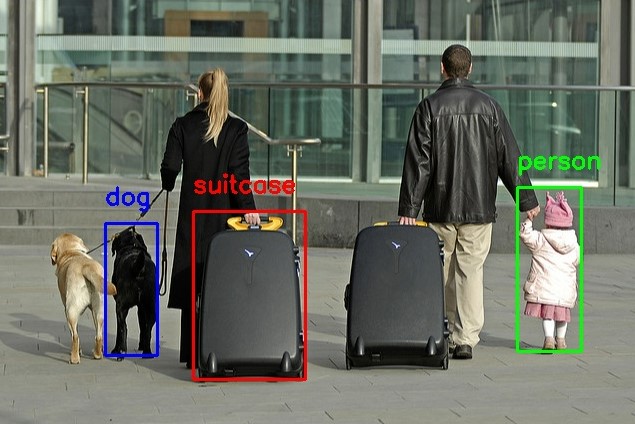}
        \includegraphics[width=\fourcolwidth\linewidth]{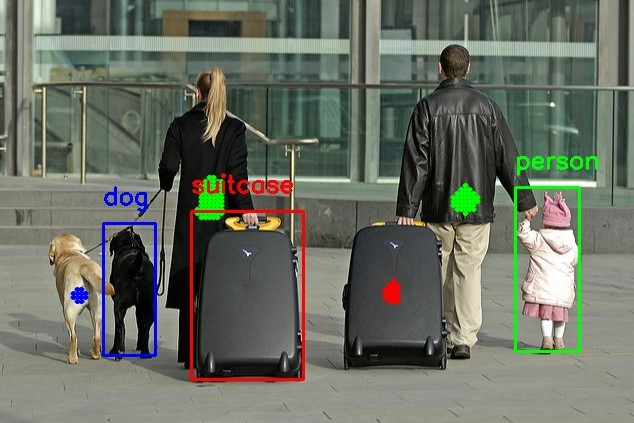}
        \includegraphics[width=\fourcolwidth\linewidth]{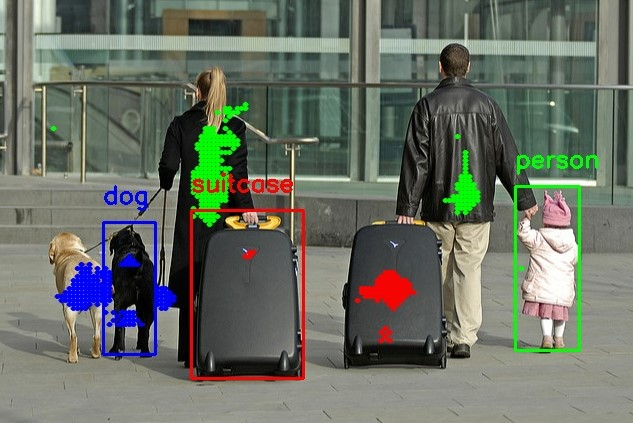} \\
        \begin{minipage}{\fourcolwidth\linewidth}
            \centering
            \figtext{(a) SPLG$@\mathbb{S}_8$}
        \end{minipage}
        \begin{minipage}{\fourcolwidth\linewidth}
            \centering
            \figtext{(b) SPLG$@\mathbb{S}_9$}
        \end{minipage}
        \begin{minipage}{\fourcolwidth\linewidth}
            \centering
            \figtext{(c) PGCL}
        \end{minipage}
        \begin{minipage}{\fourcolwidth\linewidth}
            \centering
            \figtext{(d) SPLG\_PGCL$@\mathbb{S}_9$}
        \end{minipage}
        \caption{Visualization of pseudo labels generated by SPLG(column(a),(b) and (d)) and top-$m$ positions selected by PGCL. Note that $S_i$ denotes the score threshold is $i/10$. All images are selected from the Keep1-COCO2017-Train and the preserved instances are drawn with the bounding boxes. The color of each dot denotes its according pseudo category label.}
        \label{fig:vis_hm}
\end{figure*}
% \begin{figure*}[t]
%     \begin{center}
%     \includegraphics[width=\linewidth]{Images/vis_hm.png}
%     \end{center}
%     \caption{Visualization of pseudo labels generated by SPLG and top-m position selected by PGCL.}
%     \label{fig:vis_hm}
% \end{figure*}
\begin{table*}
    \begin{center}
        \begin{tabular}{l|c|c|c|c}
            \toprule
            Task & \#instances & instances/image & instances/image/category & time(seconds) \\
            \hline
            FSOD & 36419 & 7.28 & 0.091(0.058) & 81.32   \\
            SIOD & 14674 & 2.93 & 0.037(0.004) & 38.14  \\
            \bottomrule
        \end{tabular}
    \end{center}
    \caption{Comparison of annotated cost between FSOD and SIOD task with 5000 images randomly selected from COCO2017-Train. Note that the \#instance denotes the total number of instances to be annotated. ``instances/image/category" denotes that the average number of instances for each category per image and (*) is according variance. The time cost is the average annotating time of single images.}
    \label{tab:anno_cost}
\end{table*}
In this section, we try to visualize the pseudo labels generated by the proposed Similarity-based Pseudo Label Generating module (SPLG). Note that all of positions with target values less than 1.0 are treated as penalty-reduced backgrounds as shown in main manuscript Eq.(5). We therefore visualize those high-quality positions which have large similarity with reference instances. As shown in Fig.~\ref{fig:vis_hm} SPLG$@\mathbb{S}_8$, a large number of instances are assigned pseudo class labels correctly and some instances (\eg umbrellas and birds) are ignored. However, none of instances have similarity with reference instances larger than 0.9 (SPLG$@\mathbb{S}_9$) . As for Pixel-level Group Contrastive Learning (PGCL), we select top-$m$ positions as positive samples according to self-predicted scores. As shown in Fig.~\ref{fig:vis_hm} PGCL, most of positions located at the center of unlabeled instances are selected and some instances are not selected due to the limited sampling. PGCL tends to minimize the distance between positive pairs and push away the negative pairs in embedding space, which undoubtedly facilitates mining more unlabeled instances in SPLG module. After integrated with PGCL, high-quality pseudo labels are generated with SPLG module, as shown in Fig.~\ref{fig:vis_hm} SPLG\_PGCL$@\mathbb{S}_9$. As an increasing number of unlabeled instances are mined for training, the performance of the detector is improved naturally.
\subsection{Visualization for Faster-RCNN and FCOS}
\begin{figure*}[t]
    \centering
        \includegraphics[width=0.33\linewidth]{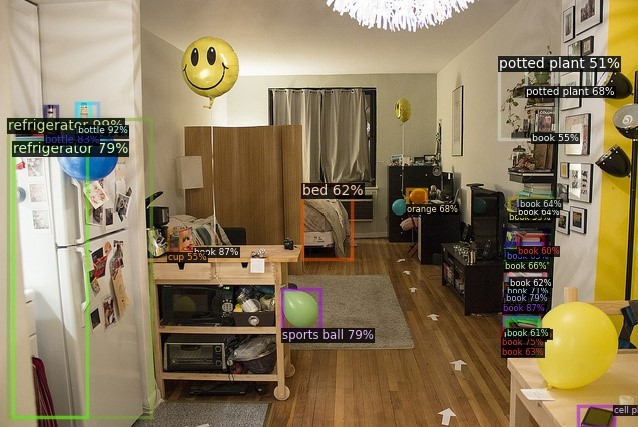} 
        \includegraphics[width=0.33\linewidth]{Images/vis_frcnn/keep1_000000014038.jpg} 
        \includegraphics[width=0.33\linewidth]{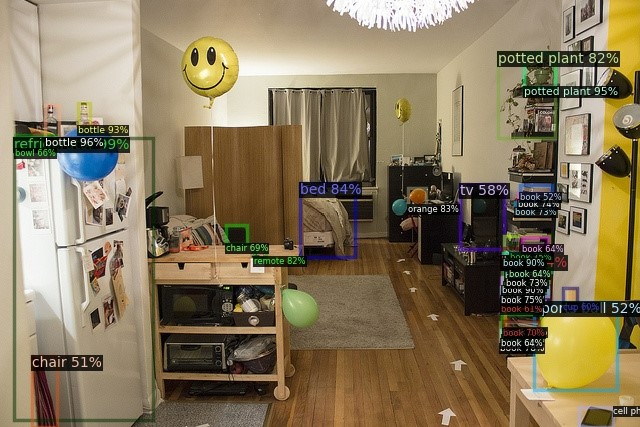}  \\
        \includegraphics[width=0.33\linewidth]{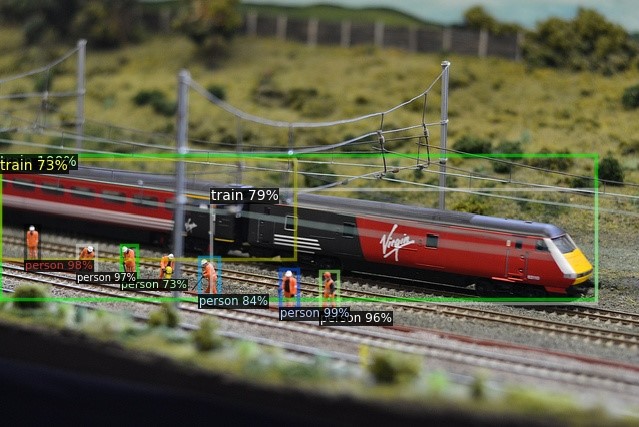} 
        \includegraphics[width=0.33\linewidth]{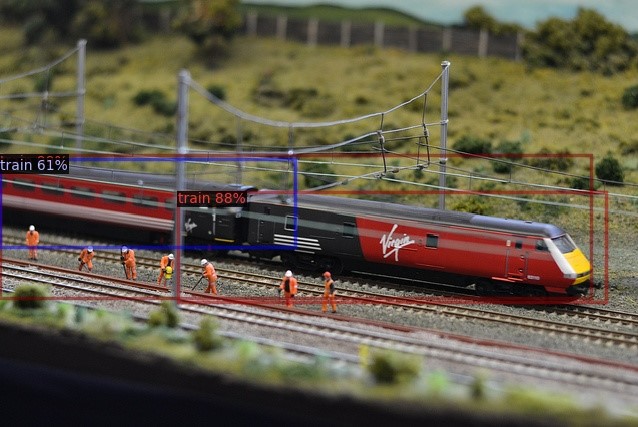} 
        \includegraphics[width=0.33\linewidth]{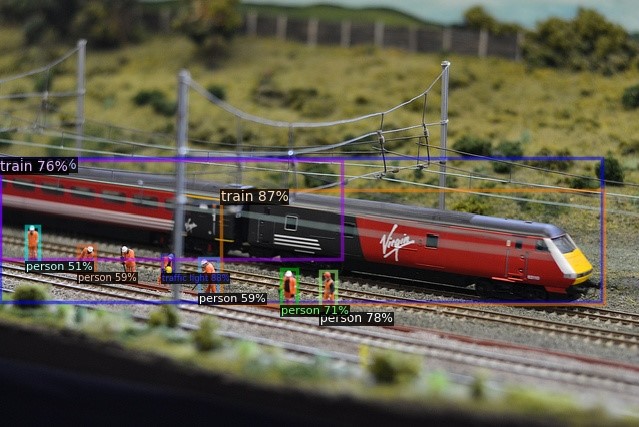}  \\
        \includegraphics[width=0.33\linewidth]{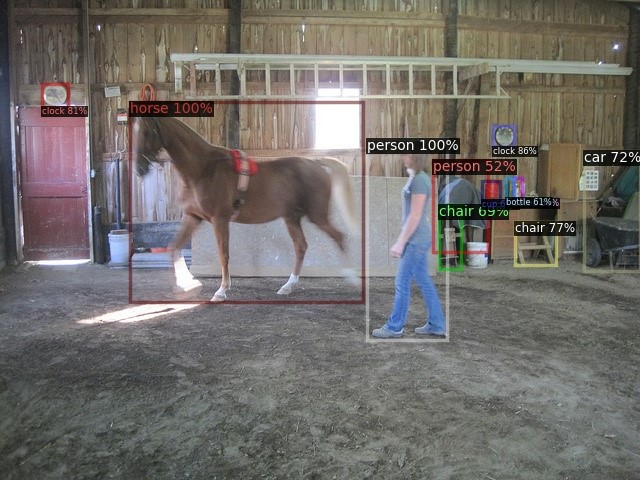} 
        \includegraphics[width=0.33\linewidth]{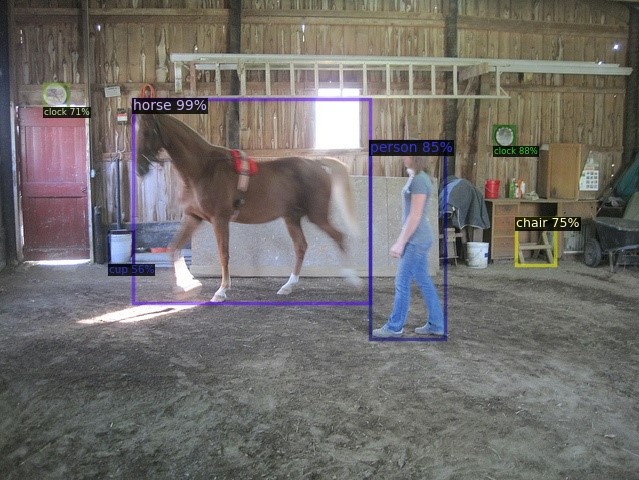} 
        \includegraphics[width=0.33\linewidth]{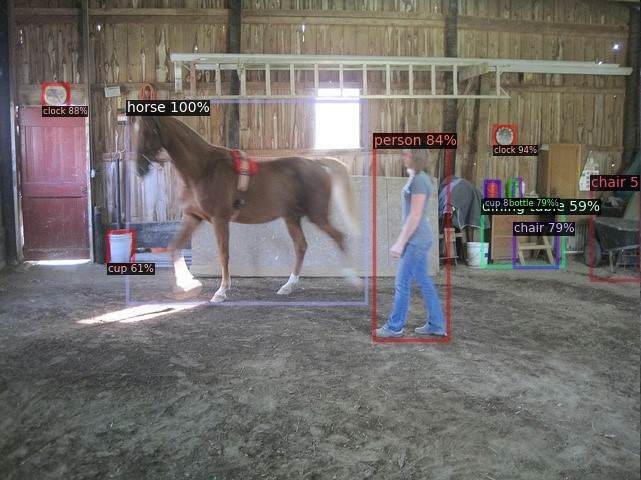} \\ 
        \begin{minipage}{0.33\linewidth}
            \centering
            \figtext{(a) FSOD}
        \end{minipage}
        \begin{minipage}{0.33\linewidth}
            \centering
            \figtext{(b) SIOD(base)}
        \end{minipage}
        \begin{minipage}{0.33\linewidth}
            \centering
            \figtext{(c) SIOD(DMiner)}
        \end{minipage}
        \caption{Visualization of Faster-RCNN-Res50-C4 on different tasks with score threshold 0.5. Note that SIOD(base) denotes directly training the detector on SIOD task and SIOD(DMiner) denotes that the detector is equipped with DMiner.}
        \label{fig:vis_frcnn}
\end{figure*}
% \begin{figure*}[h]
%     \begin{center}
%     \includegraphics[width=\linewidth]{Images/vis_frcnn.png}
%     \end{center}
%     \caption{Visualization of Faster-RCNN-Res50-C4 on different tasks with score threshold 0.5. (a) FSOD, (b) SIOD(base), (c) SIOD(DMiner).}
%     \label{fig:vis_frcnn}
% \end{figure*}
\begin{figure*}[t]
    \centering
        \includegraphics[width=0.33\linewidth]{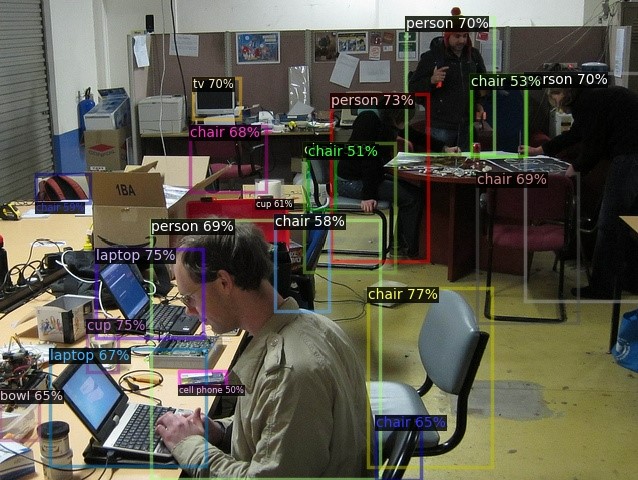} 
        \includegraphics[width=0.33\linewidth]{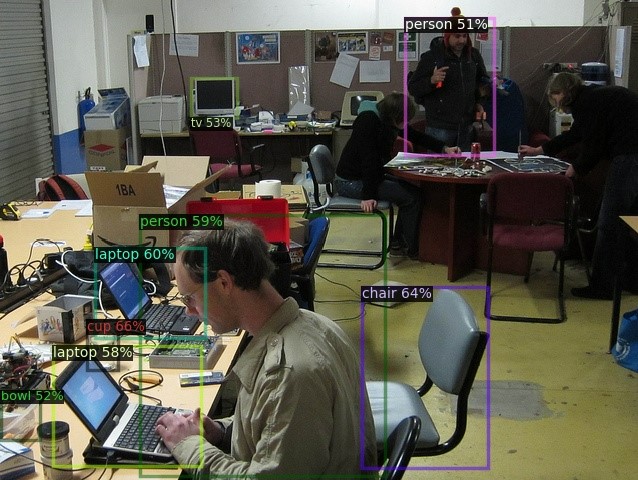} 
        \includegraphics[width=0.33\linewidth]{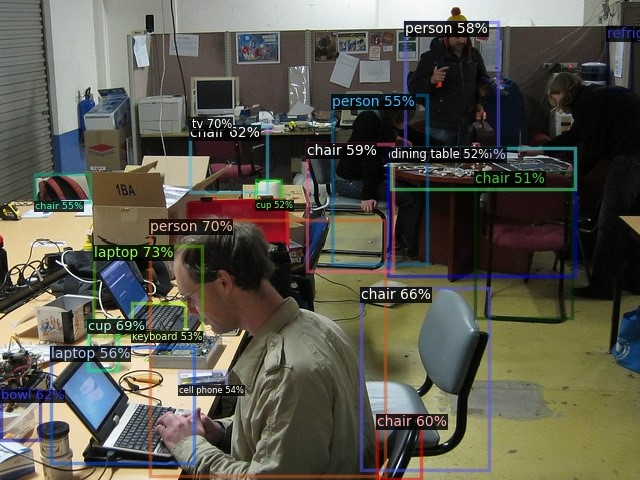}  \\
        \includegraphics[width=0.33\linewidth]{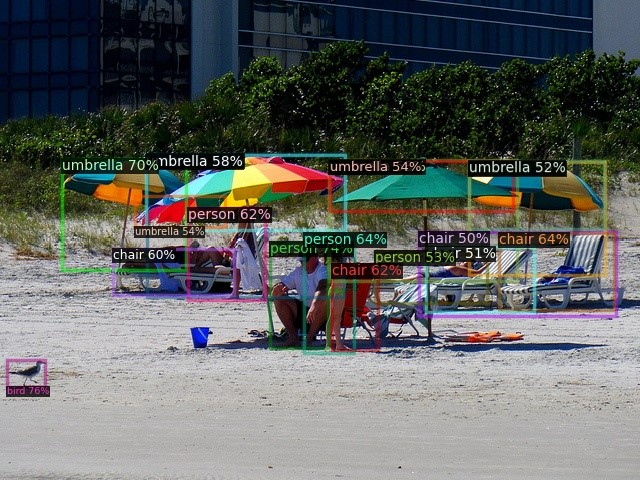} 
        \includegraphics[width=0.33\linewidth]{Images/vis_fcos/keep1_000000066038.jpg} 
        \includegraphics[width=0.33\linewidth]{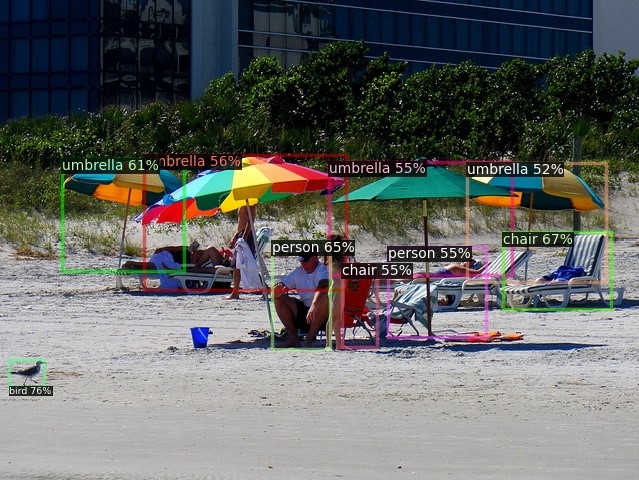}  \\
        \includegraphics[width=0.33\linewidth]{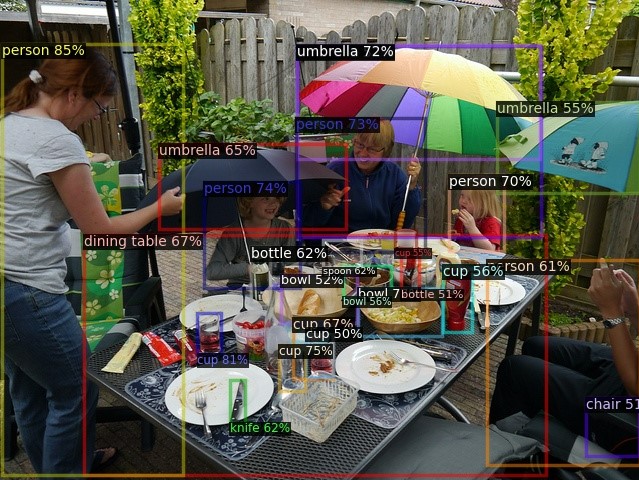} 
        \includegraphics[width=0.33\linewidth]{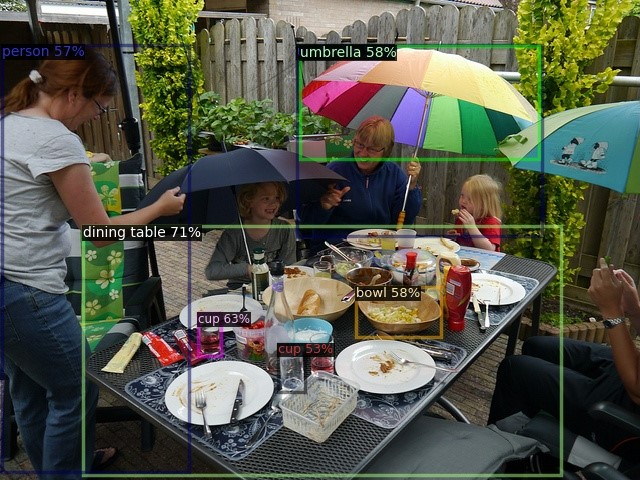} 
        \includegraphics[width=0.33\linewidth]{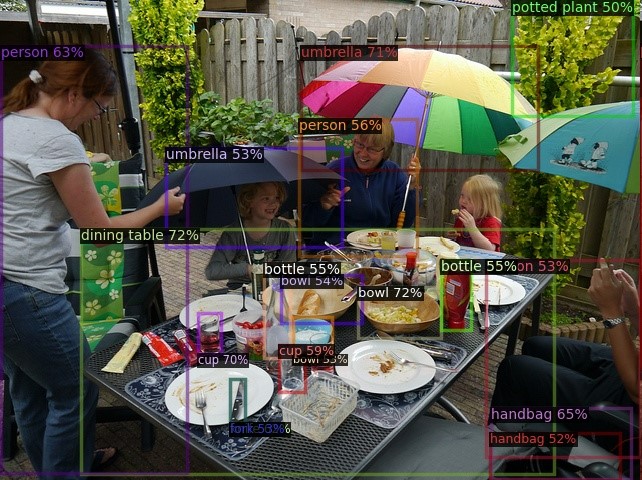} \\ 
        \begin{minipage}{0.33\linewidth}
            \centering
            \figtext{(a) FSOD}
        \end{minipage}
        \begin{minipage}{0.33\linewidth}
            \centering
            \figtext{(b) SIOD(base)}
        \end{minipage}
        \begin{minipage}{0.33\linewidth}
            \centering
            \figtext{(c) SIOD(DMiner)}
        \end{minipage}
        \caption{Visualization of FCOS-RCNN-Res50-FPN on different tasks with score threshold 0.5. Note that SIOD(base) denotes directly training the detector on SIOD task and SIOD(DMiner) denotes that the detector is equipped with DMiner.}
        \label{fig:vis_fcos}
\end{figure*}
% \begin{figure*}[h]
%     \begin{center}
%     \includegraphics[width=\linewidth]{Images/vis_fcos.png}
%     \end{center}
%     \caption{Visualization of FCOS-Res50-FPN on different tasks with score threshold 0.5. (a) FSOD, (b) SIOD(base), (c) SIOD(DMiner)}
%     \label{fig:vis_fcos}
% \end{figure*}
Both Faster-RCNN and FCOS are confronted with large performance degradation when applying them to SIOD task, since most of unlabeled instances are treated as backgrounds mistakenly. After equipped with the proposed DMiner, they achieve better performance as reported in main manuscript. In this section, we visualize the detected results for clear comparison. As shown in Fig.~\ref{fig:vis_frcnn}, SIOD(base) fails to detect those small objects (\eg books, pedestrians) while SIOD(DMiner) locates them successfully. As for FCOS, the detector equipped with DMiner also achieves obvious advance compared with SIOD(base) as shown in Fig.~\ref{fig:vis_fcos}.
\subsection{Comparison of Annotated Cost}
\begin{table*}[t]
    \begin{center}
    \begin{tabular}{c|c|c|c|c|c|c|c}
        \toprule
        category & \#instances & sample\_ratio & keep\_ratio & category & \#instances & sample\_ratio & keep\_ratio \\
        \hline
        person & 257253 & 0.042 & 0.252 & bicycle & 7056 & 0.040 & 0.474 \\ 
        car & 43533 & 0.041 & 0.289 & motorcycle & 8654 & 0.034 & 0.498 \\ 
        airplane & 5129 & 0.045 & 0.586 & bus & 6061 & 0.043 & 0.681 \\ 
        train & 4570 & 0.040 & 0.800 & truck & 9970 & 0.046 & 0.598 \\ 
        boat & 10576 & 0.045 & 0.300 & traffic light & 12842 & 0.044 & 0.330 \\ 
        fire hydrant & 1865 & 0.041 & 0.908 & stop sign & 1983 & 0.037 & 0.890 \\ 
        parking meter & 1283 & 0.018 & 0.826 & bench & 9820 & 0.041 & 0.585 \\ 
        bird & 10542 & 0.045 & 0.255 & cat & 4766 & 0.040 & 0.837 \\ 
        dog & 5500 & 0.052 & 0.704 & horse & 6567 & 0.049 & 0.443 \\ 
        sheep & 9223 & 0.048 & 0.186 & cow & 8014 & 0.053 & 0.222 \\ 
        elephant & 5484 & 0.035 & 0.424 & bear & 1294 & 0.035 & 0.733 \\ 
        zebra & 5269 & 0.042 & 0.315 & giraffe & 5128 & 0.036 & 0.505 \\ 
        backpack & 8714 & 0.040 & 0.609 & umbrella & 11265 & 0.045 & 0.340 \\ 
        handbag & 12342 & 0.042 & 0.554 & tie & 6448 & 0.037 & 0.637 \\ 
        suitcase & 6112 & 0.040 & 0.393 & frisbee & 2681 & 0.035 & 0.926 \\ 
        skis & 6623 & 0.038 & 0.472 & snowboard & 2681 & 0.037 & 0.626 \\ 
        sports ball & 6299 & 0.043 & 0.725 & kite & 8802 & 0.042 & 0.287 \\ 
        baseball bat & 3273 & 0.043 & 0.810 & baseball glove & 3747 & 0.043 & 0.679 \\ 
        skateboard & 5536 & 0.053 & 0.577 & surfboard & 6095 & 0.038 & 0.611 \\ 
        tennis racket & 4807 & 0.040 & 0.782 & bottle & 24070 & 0.043 & 0.354 \\ 
        wine glass & 7839 & 0.044 & 0.314 & cup & 20574 & 0.046 & 0.429 \\ 
        fork & 5474 & 0.045 & 0.587 & knife & 7760 & 0.049 & 0.524 \\ 
        spoon & 6159 & 0.045 & 0.564 & bowl & 14323 & 0.044 & 0.524 \\ 
        banana & 9195 & 0.049 & 0.245 & apple & 5776 & 0.035 & 0.325 \\ 
        sandwich & 4356 & 0.045 & 0.526 & orange & 6302 & 0.034 & 0.292 \\ 
        broccoli & 7261 & 0.041 & 0.271 & carrot & 7758 & 0.039 & 0.237 \\ 
        hot dog & 2884 & 0.037 & 0.444 & pizza & 5807 & 0.041 & 0.540 \\ 
        donut & 7005 & 0.047 & 0.212 & cake & 6296 & 0.030 & 0.516 \\ 
        chair & 38073 & 0.050 & 0.300 & couch & 5779 & 0.042 & 0.736 \\ 
        potted plant & 8631 & 0.053 & 0.479 & bed & 4192 & 0.036 & 0.854 \\ 
        dining table & 15695 & 0.046 & 0.719 & toilet & 4149 & 0.041 & 0.859 \\ 
        tv & 5803 & 0.044 & 0.795 & laptop & 4960 & 0.040 & 0.719 \\ 
        mouse & 2261 & 0.046 & 0.790 & remote & 5700 & 0.045 & 0.521 \\ 
        keyboard & 2854 & 0.052 & 0.728 & cell phone & 6422 & 0.045 & 0.691 \\ 
        microwave & 1672 & 0.035 & 0.966 & oven & 3334 & 0.046 & 0.890 \\ 
        toaster & 225 & 0.040 & 0.889 & sink & 5609 & 0.044 & 0.829 \\ 
        refrigerator & 2634 & 0.045 & 0.915 & book & 24077 & 0.041 & 0.227 \\ 
        clock & 6320 & 0.048 & 0.721 & vase & 6577 & 0.038 & 0.553 \\ 
        scissors & 1464 & 0.045 & 0.576 & teddy bear & 4729 & 0.032 & 0.523 \\ 
        hair drier & 198 & 0.061 & 1.000 & toothbrush & 1945 & 0.047 & 0.418 \\ 
        \bottomrule
    \end{tabular}
    \end{center}
    \caption{The detailed information of 5000 sampled images. \#instances is the total number of instances for each category in COCO2017-Train. The sample\_ratio denotes the proportion of instances w.r.t \#instances in 5000 sampled images and the average of sample\_ratio is 0.04, which is nearly same as the sampling ratio(5000/117316). The keep\_ratio denotes the proportion of instances to be annotated in 5000 sampled images under the SIOD setup and the average of keep\_ratio is 0.57.}
    \label{tab:instances_details}
\end{table*}
Although  about 60\% instance annotations are reduced under the SIOD setup compared with FSOD on COCO2017, it is still unable to directly reflect the difficulty of annotating instances between SIOD and FSOD task. We therefore conduct a practical annotating experiment to obtain real statistics of annotated cost. We first randomly select 5000 images from COCO2017-Train. The detailed information is reported in Table~\ref{tab:instances_details}. Then six professional female annotators are divided into two groups. One is asked to annotate all instances for FSOD task and another is asked to annotate one instance for each existing category in each image for SIOD task. Note that the average age of them is about 23.  Additionally, they annotate the whole samples independently and we finally compute the average annotating time among the group for each task. As shown in Table~\ref{tab:anno_cost}, only about 40\% (14674/36419) instances are needed to be annotated in 5000 sampled images for SIOD task, which is consistent  with the whole dataset. More specifically, it reduces about 53.1\% annotating time per image under the SIOD setup compared with FSOD, which demonstrates that SIOD setup has large potential to practically reduce the annotated cost for object detection. 
\end{document}